%% file: main.tex
\documentclass{article} 
\ifdefined\pdfvariable
  \pdfvariable minorversion=7
\else
\fi
\PassOptionsToPackage{table}{xcolor}
\usepackage{iclr2027_conference,times}
\usepackage{tcolorbox}
\input{math_commands.tex}

\usepackage{hyperref}
\usepackage{url}
\usepackage[table]{xcolor}

\usepackage[T1]{fontenc}
\usepackage{amsmath,amssymb,amsfonts}
\usepackage{mathrsfs}
\usepackage{graphicx}
\usepackage{booktabs}
\usepackage{array}
\usepackage{multirow}
\usepackage{makecell}
\usepackage{subcaption}
\usepackage{caption}
\usepackage{float}
\usepackage{wrapfig}
\usepackage{enumitem}
\usepackage{tikz}
\usepackage[ruled,vlined]{algorithm2e}
\usepackage[capitalize]{cleveref}


\newcommand{\perf}[3]{\makecell{#1 \scriptsize $\pm$ #2 \\[-0.5ex] \scriptsize(#3\%$\downarrow$)}}
\newcommand{\var}[2]{#1 \scriptsize $\pm$ #2}
\newcommand{\perfs}[4]{\makecell{#1 \scriptsize $\pm$ #2 \\[-0.5ex] \scriptsize(#3\%$#4$)}}

\newcommand{\Reffig}[1]{\cref{#1}}
\newcommand{\Refsec}[1]{\cref{#1}}
\newcommand{\Refalg}[1]{\cref{#1}}

\newcommand{\Reftab}[1]{\cref{#1}}

\def\MethodName{\textit{GEM3D}~}
\def\MethodNameNs{\textit{GEM3D}}
\def\PretrainingName{\textit{GEM3D Pretraining}~}
\def\PolicyName{\textit{GEM3D Policy}~}
\newcommand{\myblue}[1]{\textcolor{blue}{#1}}

\definecolor{gray}{gray}{0.5}
\definecolor{questionboxbg}{HTML}{F5F5F5}
\iclrfinalcopy 
\begin{document}
\vspace*{-8mm}
\raggedbottom
\title{Learning Geometrically-Grounded Amodal 3D Representations
for View-Generalizable Robotic Manipulation}

\author{
\textbf{Di Zhang}$^{1*}$\quad
\textbf{Weicheng Duan}$^{2*}$\quad
\textbf{Dasen Gu}$^{2}$\quad
\textbf{Hongye Lu}$^{2}$\quad
\textbf{Hai Zhang}$^{3}$ \\
\textbf{Hang Yu}$^{2}$\quad
\textbf{Guang Chen}$^{2}$\quad
\textbf{Chuan Wen}$^{1\dagger}$\quad
\textbf{Junqiao Zhao}$^{2\dagger}$ \\[1.5mm]
$^{1}$Shanghai Jiao Tong University \quad $^{2}$Tongji University \quad $^{3}$The University of Hong Kong \\
}

\maketitle
\lhead{Preprint}
\begingroup
\renewcommand{\thefootnote}{\fnsymbol{footnote}}
\footnotetext[1]{\raggedright Equal contribution: \texttt{zhangdi\_dino@sjtu.edu.cn}, \texttt{2252109@tongji.edu.cn}.}
\footnotetext[2]{\raggedright Corresponding authors: \texttt{wenchuan@sjtu.edu.cn}, \texttt{zhaojunqiao@tongji.edu.cn}.}
\endgroup
\vspace{-6mm}

\input{sec/0_abstract}   
\input{sec/teaser.tex}
\input{sec/1_intro}
\input{sec/2_related_work}
\input{sec/3_method}
\input{sec/4_experiments}
\input{sec/5_discussion}
\input{sec/5_conclusion}

\clearpage
\section*{LLM Usage Statement}
We used large language models (LLMs) to assist with language polishing, grammar correction, and improvements to the clarity, readability, and flow of the manuscript. The authors reviewed and revised the LLM-assisted text and take full responsibility for the content and accuracy of the final manuscript.

\bibliography{main}
\bibliographystyle{iclr2027_conference}

\input{sec/X_suppl}

\end{document}

%% file: math_commands.tex
\usepackage{amsmath,amsfonts,bm}

\def\eqref#1{equation~\ref{#1}}

\def\1{\bm{1}}

\DeclareMathAlphabet{\mathsfit}{\encodingdefault}{\sfdefault}{m}{sl}
\SetMathAlphabet{\mathsfit}{bold}{\encodingdefault}{\sfdefault}{bx}{n}



%% file: sec/0_abstract.tex
\begin{abstract}
Real-world robotic manipulation requires visuomotor policies capable of robust 3D scene reasoning under varying camera viewpoints.
While recent 3D-aware manipulation policies have shown promise, they still face several limitations:
(i) reliance on multi-view observations during inference, which is impractical in camera-constrained deployments;
(ii) insufficient geometric fidelity in learned representations, limiting precise and viewpoint-robust control; and
(iii) the lack of effective mechanisms for transferring pretrained 3D representations to downstream visuomotor policies.
To address these challenges, we present \textbf{\MethodName} (\myblue{Ge}o\myblue{m}etrically-Grounded \myblue{3D} Manipulation), a unified 3D representation and policy learning framework for view-generalizable robotic manipulation.
\MethodName learns amodal 3D representations from single-view RGB-D observations by jointly enforcing point-cloud reconstruction and Gaussian-splatting-based novel-view consistency during pretraining.
A multi-step distillation strategy then transfers the learned geometric understanding into a deployable single-view visuomotor policy for downstream manipulation control.
Experiments on 12 RLBench tasks and three real-world tasks demonstrate strong manipulation performance. Further zero-shot evaluations on eight RLBench tasks and the real-world tasks show robust generalization to unseen camera poses and substantial gains over existing methods.
\end{abstract}


%% file: sec/teaser.tex
{
\setlength{\parskip}{0pt}
\setlength{\parindent}{0pt}

\setlength{\abovedisplayskip}{0pt}
\setlength{\belowdisplayskip}{0pt}
\setlength{\textfloatsep}{0pt}
\setlength{\intextsep}{0pt}
\setlength{\floatsep}{0pt}

\noindent
\vbox{%
{\rmfamily\fontsize{9pt}{11pt}\selectfont\textbf{(I) GEM3D Pretraining: Partial-to-Holistic Scene Completion
}}\par
\includegraphics[
    width=\textwidth,
    trim=-1 0 1 0,
    clip
]{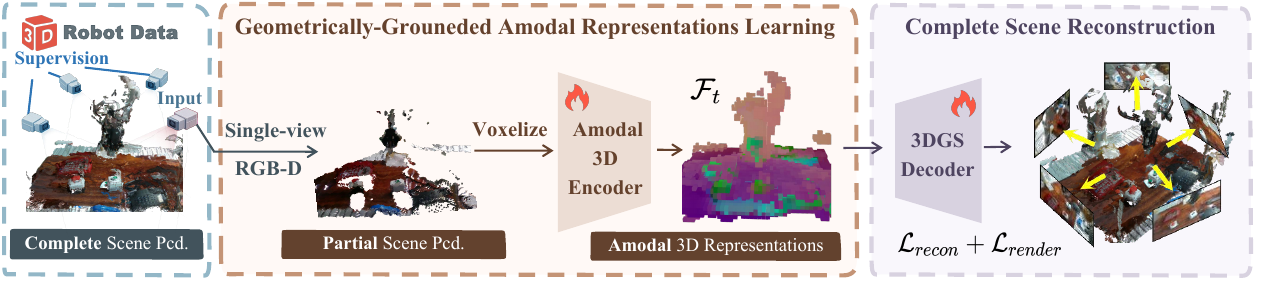}
}

\noindent
\begingroup
\setlength{\parskip}{0pt}
\setlength{\baselineskip}{0pt}

\vtop{%
\hsize=\linewidth

\noindent
{\rmfamily\fontsize{9pt}{11pt}\selectfont\textbf{(II) GEM3D Policy: Multi-step 3D Representation Distillation
}}\par

\vskip 2pt

\noindent
\begin{minipage}[t]{0.69\linewidth}
\includegraphics[
    width=\linewidth,
    trim=0 0 0 0,
    clip
]{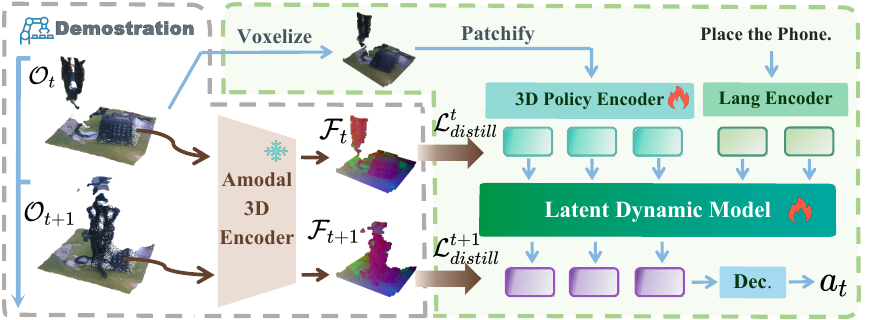}
\end{minipage}%
\hspace*{0.015\linewidth}%
\begin{minipage}[t]{0.31\linewidth}
\includegraphics[
    width=\linewidth,
    trim=7 0 6 0,
    clip
]{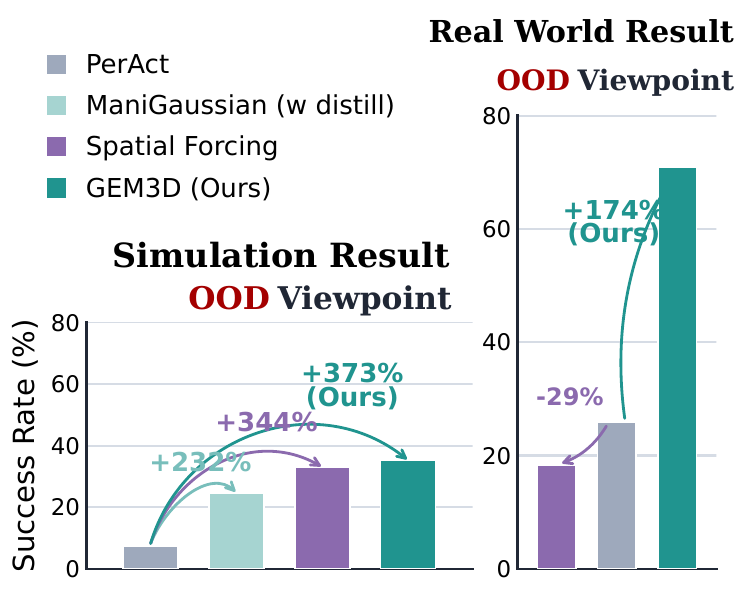}
\end{minipage}

}%
\endgroup

\captionof{figure}{
    \textbf{
    Overview of \MethodNameNs.}
    \MethodName comprises two key components:
    (1) \textit{GEM3D Pretraining}, which learns amodal 3D representations via auxiliary scene reconstruction tasks;
    and (2) \textit{GEM3D Policy}, which distills pretrained 3D representations into a visuomotor policy for view-generalizable manipulation.
}
\label{fig:teaser}
}

%% file: sec/1_intro.tex
\section{Introduction}
\label{sec:intro}
\vspace{-0.75em}
\noindent
\begingroup
\setlength{\fboxsep}{3pt}
\setlength{\fboxrule}{0.6pt}
\raisebox{-7pt}{\fcolorbox{black}{questionboxbg}{%
  \parbox{\dimexpr\linewidth-2\fboxsep-2\fboxrule\relax}{%
\textbf{Amodal}~\citep{wu2025amodal3r, ao2023image, zhou2025amodalgen3d}
refers to the ability to infer the complete spatial structure of a scene
from partial observations, such as a single-view input. In real-world
manipulation, robots often act under partial observability caused by
occlusions and limited viewpoints. We therefore argue that learning
amodal representations is essential for building effective 3D policies.%
  }%
}}%
\endgroup
\par
Learning end-to-end visuomotor policies enables robots to perceive their surroundings and act upon visual understanding~\citep{chiDiffusionPolicyVisuomotor2024,kimOpenVLAOpenSourceVisionLanguageAction2024,chiUniversal2024,hansenTDMPC2ScalableRobust2024,zhang2024,haWorldModels2018,hafnerMasteringAtariDiscrete2020}.
However, real-world environments are often spatially complex and partially occluded, and camera viewpoints may vary across deployments.
Robust manipulation therefore requires policies that can infer amodal 3D scene structure from partial observations and make decisions that generalize well across viewpoints.

\vspace{-0.5em}
Most existing visuomotor policies rely on 2D visual encoders that compress images into latent states~\citep{yaratsImprovingSampleEfficiency2020,laskinCURLContrastiveUnsupervised2020,hafnerDREAMCONTROLLEARNING2020}.
To improve spatial reasoning, recent methods leverage multi-view inputs to learn 3D-aware visual representations~\citep{goyal2023,qian3DMVP2025,shridhar2023perceiver}.
However, their reliance on synchronized multi-view observations makes deployment difficult in practical robotic settings with limited camera availability.
To reduce this dependency, some single-view 3D policies incorporate reconstruction objectives, such as Neural Radiance Fields (NeRFs)~\citep{mildenhall2022} or Gaussian Splatting~\citep{kerbl2023}, to obtain 3D-aware perception~\citep{driessReinforcementLearningNeurala,wangReinforcementLearningGeneralizable2024,luManiGaussianDynamicGaussian2024,ze2023,ze3DDiffusionPolicy2024}.
Nevertheless, these methods mainly reconstruct geometry directly observable from the input view and often struggle to infer occluded scene structures.

To address these limitations, we propose \textbf{\MethodName} (\myblue{Ge}o\myblue{m}etrically-Grounded \myblue{3D} Manipulation), a two-stage 3D representation-policy learning framework for view-generalizable robotic manipulation. 
In the first stage, \MethodName learns amodal 3D representations from single-view RGB-D inputs by using multi-view observations only as geometric supervision. 
Given a partial observation from one camera, the model is trained to reconstruct the complete scene point cloud, and enforce view-consistent geometric reasoning through Gaussian-splatting-based novel-view supervision. 
Such a representation is well suited for manipulation, as it enables the robot to infer the scene structure from any single camera observation, including occluded task-relevant objects and structures.

In the second stage, we transfer the learned spatial understanding to a single-view visuomotor policy through a multi-step distillation paradigm. 
Directly finetuning a pretrained 3D encoder with action supervision can easily degrade its geometric reasoning capability.
To preserve the learned spatial understanding, we freeze the pretrained representation model as a geometric teacher and distill its amodal scene representations into a lightweight policy encoder.
We therefore use task execution from unseen camera viewpoints as a practical measure of whether pretrained 3D knowledge is preserved in the policy.
As a result, even when trained only on fixed single-view demonstrations, \MethodName can perform manipulation from out-of-distribution camera viewpoints at inference time.

We evaluate \MethodName on three real-world tasks and 12 RLBench tasks, achieving 44.2\% average in-domain success on the latter, 9.1 points above the strongest baseline.
We further test view generalization on eight RLBench tasks, reaching 41.0\%/29.5\% under moderate/large shifts, 11.7/9.4 points above the strongest baselines.
In the real world, \MethodName reaches 88.3\% at the training view and 78.3\%/63.3\% under moderate/large shifts, exceeding the strongest baselines by 43.3, 41.7, and 48.3 points, respectively.

In summary, the contributions of \MethodName are threefold:
(i) We propose a single-view amodal 3D representation learning framework for view-generalizable robotic manipulation that infers occlusion-aware scene geometry from partial RGB-D observations.
(ii) We introduce a geometry-preserving distillation strategy that transfers pretrained spatial reasoning capability into a lightweight deployable visuomotor policy.
(iii) We demonstrate strong manipulation performance on 12 RLBench tasks, with further viewpoint-generalization evaluations on eight RLBench tasks and three real-world tasks.

%% file: sec/2_related_work.tex
\section{Related Work}
\label{sec:related_work}

\textbf{3D Representation Learning.}
Recent visuomotor policy learning has explored diverse visual representations, from 2D image features~\citep{chiDiffusionPolicyVisuomotor2024,z2026sp} to explicit 3D structures such as voxel grids~\citep{shridhar2023perceiver} and point clouds~\citep{chen2023polarnet,guhur2023instruction,ze3DDiffusionPolicy2024,chen2026dexxlearningvisualtactiledexterous}.
Among these, novel-view synthesis (NVS) has emerged as a useful auxiliary objective for learning 3D-aware representations, as rendering unseen viewpoints encourages the model to infer scene geometry beyond the input observation~\citep{mildenhall2022,kerbl2023,poole2022dreamfusion,liu2023zero,shi2023zero123++}.
However, accurate and generalizable reconstruction remains challenging: classical NeRFs~\citep{mildenhall2022} and Gaussian Splatting~\citep{kerbl2023} often degrade outside the training distribution, and feed-forward Gaussian Splatting methods~\citep{szymanowicz2024splatter,ren2024scube,charatan2024pixelsplat}, while more generalizable, still exhibit degraded geometry and texture in dynamic robotic scenes.
These limitations highlight the need for an NVS framework that achieves both high-fidelity reconstruction and robust generalization across diverse robotic scenes.

\textbf{Visuomotor Policy Representation  Learning.}
Auxiliary representation learning for control mainly follows pretrain-finetune or joint-training paradigms.
Pretrain-finetune methods initialize visual encoders from large-scale 2D or 3D perception datasets before adapting them to downstream control~\citep{xiao2022masked,wang2022vrl3,yan2024dnact,ze2023visual,qian3DMVP2025,wang2026r2sevalrobotevaluationrealtosim}, but domain discrepancies between perception pretraining and policy learning can cause suboptimal transfer or even degrade control performance~\citep{hansen2022pre,hou20254d}.
Joint-training methods optimize visuomotor policies together with auxiliary objectives such as NVS~\citep{ze2023,luManiGaussianDynamicGaussian2024}, yet the competing objectives can lead to imbalanced optimization, where the learned representation remains weak and the policy still overfits to training viewpoints.
In contrast to prior works, we propose a multi-step distillation framework that softly guides policy encoder training with pretrained 3D representations, effectively preserving geometric knowledge acquired during pretraining.

%% file: sec/3_method.tex
\section{Method}
\label{sec:method}

\subsection{Problem Formulation}

We consider language-conditioned robotic manipulation from a single RGB-D camera. At each timestep $t$, the robot receives $o_t^v=\{I_t^v,D_t^v,K_t^v\}$, where $v$ is the camera viewpoint and $I_t^v$, $D_t^v$, and $K_t^v$ are the RGB image, depth map, and camera parameters. Given a language instruction $l$, our goal is to learn a policy $\pi(a_t|o_t^v,l)$ that predicts action $a_t$ from an arbitrary-viewpoint observation.

\textbf{Pretraining phase.}
GEM3D learns a 3D representation $F_t$ from a multi-view dataset.
Given a single-view input $o_t^v$, the remaining views $v' \neq v$ at the same timestep provide cross-view supervision.
The goal is to make $F_t$ infer amodal 3D information from the partial observation.

\textbf{Policy learning phase.}
We then train a 3D visuomotor policy on single-view datasets with a separate policy encoder.
This encoder maps $o_t^v$ into a policy representation $x_t$, which is used to predict the action $a_t$.
The goal is to transfer the 3D understanding pretrained in $F_t$ into $x_t$, enabling geometry-aware action prediction from single-view inputs.

\textbf{Inference.}
We evaluate zero-shot view generalization as a practical proxy for 3D scene understanding by testing each episode from a random camera viewpoint unseen during policy training.

\subsection{Geometrically-Grounded Pretraining}
\label{sec:pretraining}

\begin{figure}[!t]
\centering
\includegraphics[width=\linewidth]{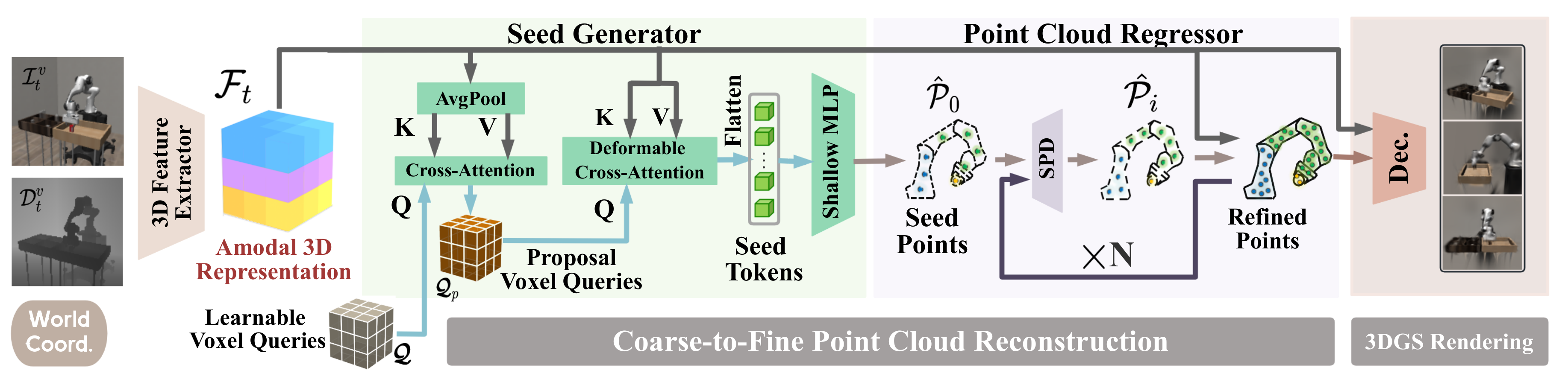}
\caption{
{
\textbf{\MethodNameNs\ 3D Pretraining Pipeline.}
Given single-view RGB-D volumetric features, we (i) first capture the coarse scene layout and local geometric details via two-stage attention, (ii) use Snowflake Point Deconvolution (SPD) blocks to progressively double the point cloud, and (iii) apply Gaussian-splatting-based novel-view rendering for fine-grained texture supervision.
}
}
\label{fig:pretrain2}
\end{figure}

\MethodNameNs's pretraining pipeline consists of three main components:
(i) a \textbf{\textit{3D Feature Extraction}} module that encodes single-view RGB-D observations into \textit{Amodal 3D Representations};
(ii) a \textbf{\textit{Point Cloud Reconstruction}} objective that learns amodal 3D scene geometry; and
(iii) a \textbf{\textit{Feed-forward 3D Gaussian Splatting}} module that captures fine-grained texture through multi-view supervision.
The detailed pipeline is illustrated in \Reffig{fig:pretrain2}.

\textbf{3D Feature Extraction.}
As shown in \Reffig{fig:teaser}, we represent the scene with a voxel-based 3D feature $F_t \in \mathbb{R}^{D^3 \times C}$ encoded from a single-view RGB-D observation $o_t^v=\{I_t^v,D_t^v,K_t^v\}$.
Voxels provide a structured 3D representation in which each token corresponds to a local spatial region, making them well suited for preserving geometric layout.
While the input RGB-D view only covers part of the scene, GEM3D pretraining encourages $F_t$ to capture amodal spatial information across viewpoints; we therefore refer to $F_t$ as an \emph{Amodal 3D Representation}.
The observation is lifted into 3D, voxelized into occupancy and semantic feature volumes, and encoded into $F_t$ by a 3D U-Net encoder~\citep{cciccek20163d}; further implementation details are provided in \Refsec{sec:3d_feature_extractor_appendix}.

\textbf{Coarse-to-Fine Point Cloud Reconstruction}
To supervise amodal scene geometry, we sample all remaining viewpoints $v' \neq v$, back-project their depth maps into the world coordinate frame, and merge the resulting points into a complete point cloud $P_{\mathrm{full}}$.

As shown in \Reffig{fig:pretrain2}
Given $F_t$, GEM3D predicts the multi-view-complete point cloud in a coarse-to-fine manner, as shown in \Reffig{fig:pretrain2}.
Learnable voxel queries $Q \in \mathbb{R}^{d^3 \times C}$ first extract seed tokens from $F_t$ through a two-stage attention design: standard cross-attention captures the coarse scene layout from a downsampled feature, while 3D deformable cross-attention refines the tokens with local geometric details from the full-resolution feature.
The refined seed tokens are decoded into seed points $\hat{P}_0 \in \mathbb{R}^{d^3 \times 3}$, which are then progressively densified by Snowflake Point Deconvolution (SPD) blocks~\citep{xiang2021snowflakenet}.
At each refinement stage, each point generates $r$ child points, expanding $\hat{P}_{i-1} \in \mathbb{R}^{r^{i-1}d^3 \times 3}$ to $\hat{P}_i \in \mathbb{R}^{r^i d^3 \times 3}$ and yielding multi-scale predictions $\{\hat{P}_i\}$ for reconstruction supervision. Implementation details are provided in \Refsec{sec:coarse2fine_appendix}.

At each refinement stage, we randomly sample a subset $P_i \subset P_{\mathrm{full}}$ with the same size as $\hat{P}_i$ and supervise the prediction with a symmetric Chamfer distance:
\begin{equation}
L_{\mathrm{rec}}=\sum_i \left(
\frac{1}{|\hat{P}_i|}\sum_{\hat{p}\in \hat{P}_i}\min_{p\in P_i}\|\hat{p}-p\|_2
+
\frac{1}{|P_i|}\sum_{p\in P_i}\min_{\hat{p}\in \hat{P}_i}\|p-\hat{p}\|_2
\right).
\end{equation}

\textbf{Feed-forward 3D Gaussian Splatting.}
The reconstructed point cloud $\hat{\mathcal{P}} \in \mathbb{R}^{N_p \times 3}$ provides explicit geometric anchors, while manipulation also requires fine-grained appearance cues such as object texture and boundaries.
We use $\hat{\mathcal{P}}$ as Gaussian centers $\mu$, sample features $\mathcal{F}_{\mathcal{G}} \in \mathbb{R}^{N_p \times C}$ from $\mathcal{F}_t$ at these locations, and regress the remaining Gaussian parameters with a ResNetFC network:
\begin{equation}
\{c, \sigma, r, s\} = \mathrm{ResNetFC}(\mathcal{F}_{\mathcal{G}}),
\end{equation}
where $c$, $\sigma$, $r$, and $s$ denote color, opacity, rotation, and scale, respectively.
The resulting Gaussians $\mathcal{G}=\{\mu,c,\sigma,r,s\}$ are rendered from held-out viewpoints $v' \neq v$ with a differentiable splatting renderer and supervised by the corresponding RGB observations $\mathcal{I}_{v'}^{t}$.
We use a focal loss~\citep{lin2017focal}, which assigns higher weights to hard-to-predict pixels and therefore emphasizes dynamic or visually ambiguous regions:
\begin{equation}
\mathcal{L}_{\mathrm{rend}}
= \mathrm{FocalLoss}(\hat{\mathcal{I}}_{v'}^{t}, \mathcal{I}_{v'}^{t}).
\end{equation}

The overall pretraining objective combines reconstruction and rendering supervision:
\begin{equation}
\mathcal{L}_{pretrain} = \mathcal{L}_{\text{rec}} + \lambda_{\text{rend}}\mathcal{L}_{\text{rend}}.
\end{equation}
where $\lambda_{\text{rend}}$ balances the two objectives.

\subsection{Distillation-based Policy Learning}
\label{sec:policy}

Instead of directly finetuning the pretrained 3D encoder, which may cause representation drift, we freeze it as a geometric teacher and train a separate PerceiverIO-style policy encoder~\citep{jaegle2021perceiver}.
Before policy encoding, each single-view observation $o_t^v$ is calibrated to the same world coordinate frame as pretraining. 
Then the policy encoder produces latent tokens $\mathbf{x}_t$ from $o_t^v$, while the frozen teacher output $\mathcal{F}_t$ is patchified into reference tokens $\tilde{\mathbf{x}}_t$. We align $\mathbf{x}_t$ with $\tilde{\mathbf{x}}_t$ using a cosine-similarity distillation loss to transfer the pretrained 3D knowledge to the policy.

For robotic control, each action can be interpreted as inducing a transition from the current latent state to the next latent state. We therefore factorize the policy as $p(\mathbf{x}_{t+1}|\mathbf{x}_t,l)\,p(a_t|\mathbf{x}_{t+1})$: a latent dynamics model first predicts $\mathbf{x}_{t+1}$ from $\mathbf{x}_t$ together with proprioceptive and language embeddings, and an action head then decodes $a_t$ from $\mathbf{x}_{t+1}$, as illustrated in \Reffig{fig:teaser}. This design enables multi-step distillation, which injects pretrained 3D knowledge and regularizes temporal transition modeling by aligning the current and predicted next latent states with their corresponding teacher tokens:
\begin{equation}
    \mathcal{L}_{\mathrm{multi\text{-}distill}} =
    \mathcal{L}_{\mathrm{distill}}^t+\mathcal{L}_{\mathrm{distill}}^{t+1}=
    \mathcal{L}_{\mathrm{cos}}(\mathbf{x}_{t}, \operatorname{sg}[\tilde{\mathbf{x}}_{t}])
    + 
    \mathcal{L}_{\mathrm{cos}}(\mathbf{x}_{t+1}, \operatorname{sg}[\tilde{\mathbf{x}}_{t+1}]).
\end{equation}
Note that policy learning uses only single-view datasets, and \textbf{neither training nor inference requires explicit geometric reconstruction}. We train the policy with imitation and distillation losses:
\begin{equation}
\mathcal{L}_{\mathrm{policy}} =
\mathcal{L}_{\mathrm{action}}
+ \lambda_{\mathrm{distill}} \mathcal{L}_{\mathrm{multi\text{-}distill}}.
\end{equation}
where $\mathcal{L}_{\mathrm{action}}$ supervises action prediction and $\lambda_{\text{distill}}$ balances the two objectives.

Details of the module design and loss hyperparameter settings are provided in \Refsec{sec:objectives_appendix}.


%% file: sec/4_experiments.tex
\begingroup
\setlength{\floatsep}{6pt}
\setlength{\textfloatsep}{10pt}
\setlength{\intextsep}{8pt}
\section{Experiments}
\label{sec:experiments}

\subsection{Experimental Setup}
\label{sec:experimental_setup}

\begin{figure}[!t]
    \centering
    \includegraphics[width=1\linewidth]{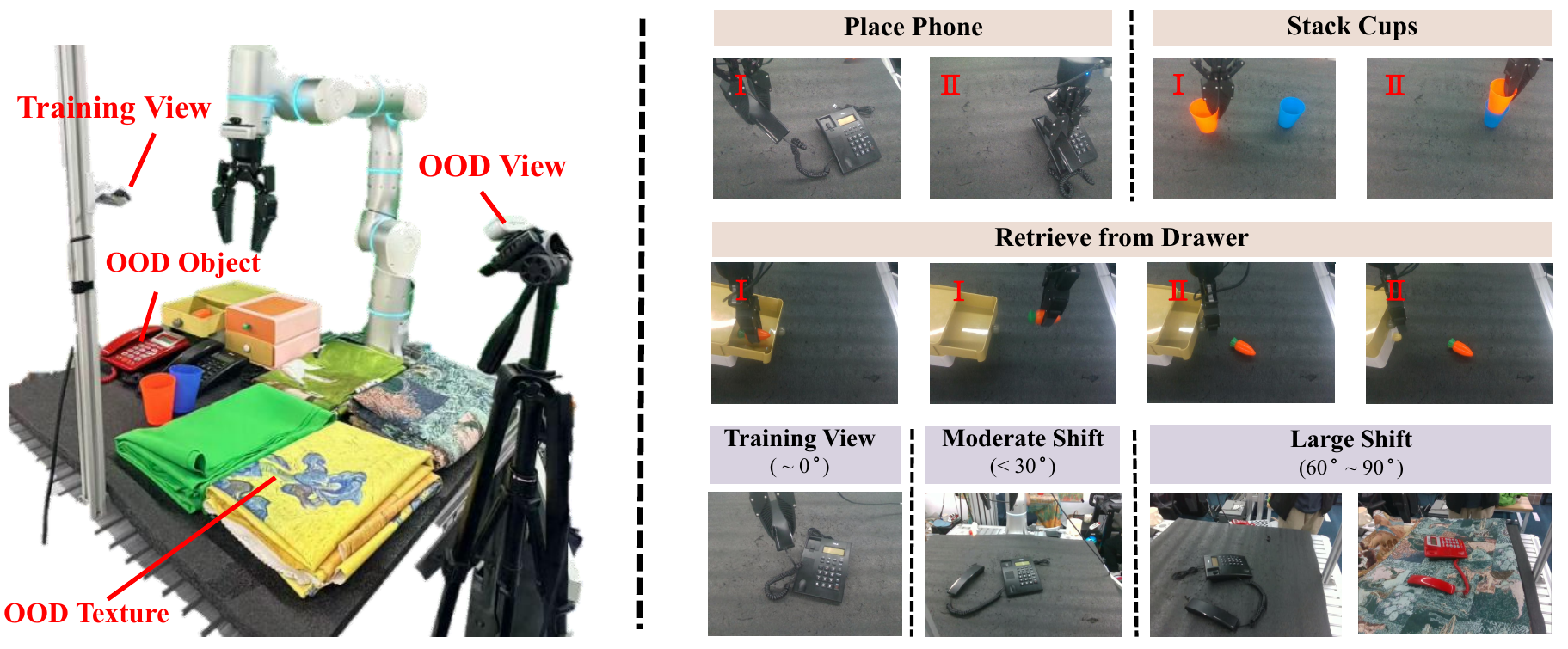}
    \caption{
        \textbf{Real-world evaluation settings.} The upper right shows the three evaluated tasks. We use two-stage scoring, with each stage marked by \textcolor{red}{I} and \textcolor{red}{II} in the figure. For example, in the \textit{Place Phone} task, correctly picking up and placing down the receiver are counted as two separate success rates. The Stage II success rate also corresponds to the full-task success rate. During evaluation, we test generalization to out-of-distribution (OOD) camera viewpoints, tabletop textures, and object colors.}
    \label{fig:setup}
\end{figure}

\textbf{Evaluation setup.}
For real-world experiments, we pretrain \MethodName on 48 RH20T~\citep{fang2023} tasks, with 7--9 viewpoints per task.
For policy learning, we choose three unseen manipulation tasks and collect 50 single-view demonstrations for each task on our Flexiv setup:
(i) \textit{Place Phone}, placing a receiver onto the phone base;
(ii) \textit{Stack Cups}, stacking one cup onto another;
and (iii) \textit{Retrieve from Drawer}, retrieving an object from a drawer and closing it.
Since the baseline methods have low full-task success rates in the real world, each task is scored with two execution stages, as detailed in \Reffig{fig:setup}.

\input{tables/sim_results}

\begin{figure}[!t]
    \centering
    \includegraphics[width=1\linewidth]{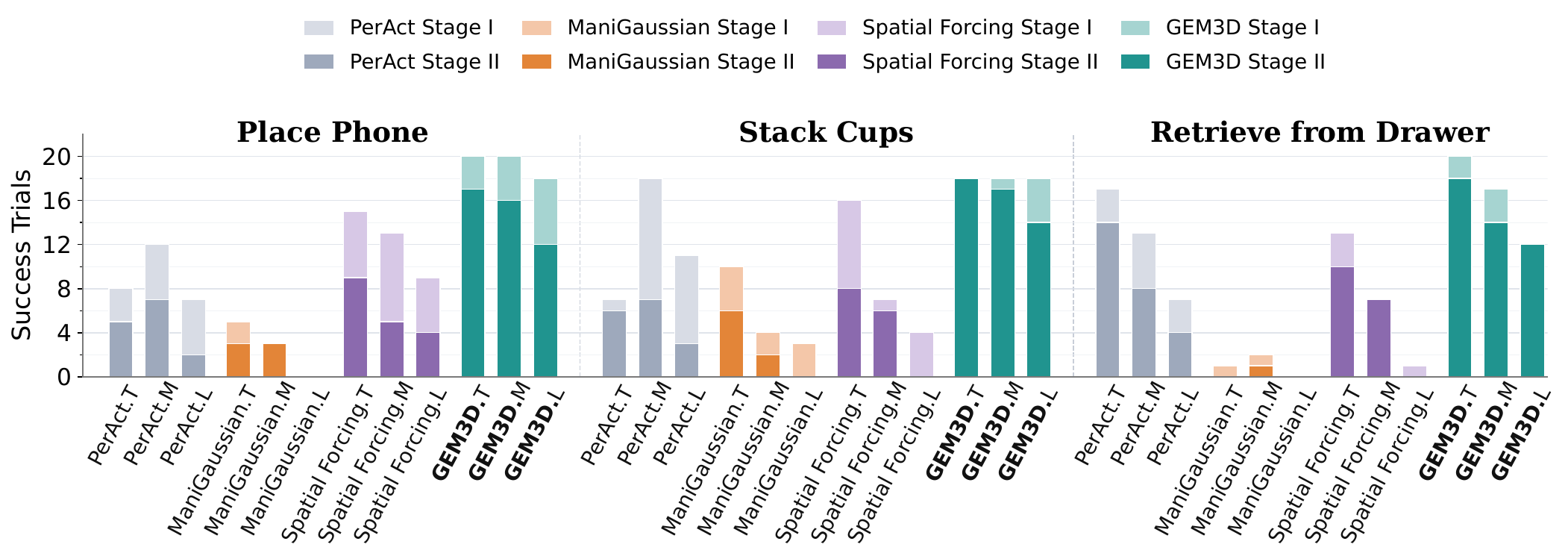}
    \caption{
        \textbf{Real-world experimental results.} We report the number of successful trials out of 20 evaluations of different stages. For each task, we evaluate three viewpoint settings: \textit{T} denotes the training view, \textit{M} denotes a moderate viewpoint shift, and \textit{L} denotes a large viewpoint shift.}
    \label{fig:real_results}
\end{figure}

In simulation, we evaluate in-domain performance on 12 RLBench~\citep{james2020rlbench} tasks across nine scenes with diverse language instructions.
During pretraining, multi-view observations are captured from eight uniformly distributed surrounding cameras.
For policy learning, each task is trained using 20 expert demonstrations collected from a fixed front-view camera.
In addition, we conduct a Robosuite~\citep{zhu2020robosuite} reinforcement learning experiment to evaluate the generality of our amodal 3D representation beyond imitation learning; details are provided in Appendix~(\ref{sec:robosuite_rl_appendix}).

\textbf{Evaluation metrics.}
To assess view generalization in both simulation and real-world settings, we conduct zero-shot evaluations under three camera configurations, as shown in \Reffig{fig:setup}: \textit{Training View}, where the camera viewpoint is the same as policy training; \textit{Moderate Shift}, where the camera viewpoint is shifted within $30^\circ$ around the robot base; and \textit{Large Shift}, where the camera viewpoint is shifted between $60^\circ$ and $90^\circ$.
For all methods, RGB-D observations from different viewpoints are calibrated into the same world coordinate frame during evaluation.
However, different viewpoint shifts induce different levels of incompleteness in the single-view point cloud, allowing us to assess each policy's implicit amodal scene inference ability.

For real-world experiments, each task-viewpoint setting is evaluated with 20 trials, for a total of $3\times 3 \times 20=180$ trials for each method.
For simulation, all models are trained with three fixed random seeds (0, 1, 2), and each seed is evaluated on 100 test episodes.
In both real-world and simulation evaluations, we randomize tabletop textures and object colors that are unseen during policy training.
On RLBench, \MethodName runs at 1.28~Hz, compared with 1.18~Hz for \textit{PerAct} and 1.36~Hz for \textit{ManiGaussian} (Appendix~\ref{sec:inference_efficiency_appendix}).

\begin{figure}[!t]
    \centering
    \includegraphics[width=0.88\linewidth]{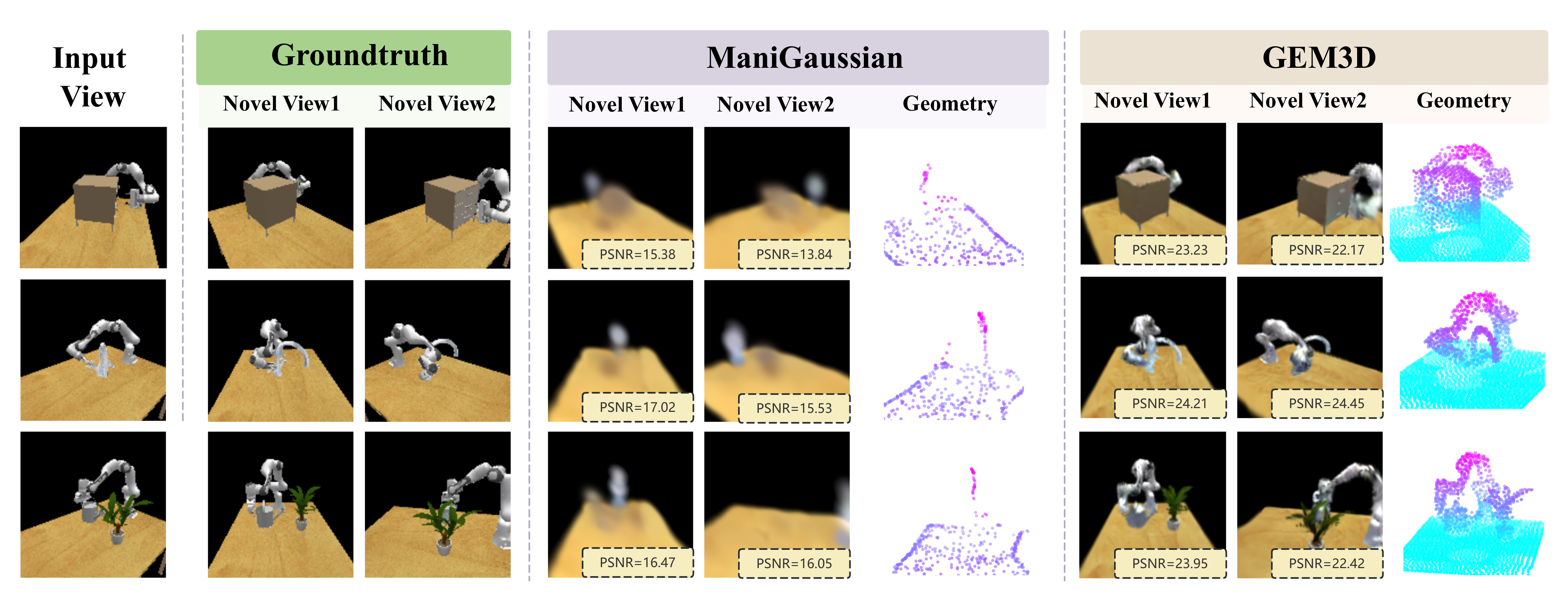}
    \caption{
        \textbf{Reconstruction quality in simulation.}
         Given a single-view input, we compare the novel-view rendering and point cloud reconstruction results of \textit{ManiGaussian} and \MethodName.}
    \label{fig:viz_sim}
\end{figure}

\textbf{Baselines.}
We first include \textbf{3D Diffusion Policy (DP3)}~\citep{ze3DDiffusionPolicy2024} as a general-purpose 3D policy baseline.
To accommodate \MethodNameNs's voxel representations, we adopt \textbf{PerceiverIO}~\citep{jaegle2021perceiver} as the policy backbone. We use \textbf{PerAct}~\citep{shridhar2023perceiver} as a basic naive baseline without auxiliary co-training, distillation, or multi-view pretraining. All the following methods, including our \MethodName, use the same PerceiverIO architecture. Other baselines include \textbf{GNFactor}~\citep{ze2023}, which uses NeRF-based auxiliary representation learning, and \textbf{ManiGaussian}~\citep{luManiGaussianDynamicGaussian2024}, which jointly trains a 3D Gaussian Splatting representation with the policy.
We further combine alternative representations with our distillation strategy. Specifically, \textbf{ManiGaussian (w/ Distill)} distills the pretrained ManiGaussian representations into the PerceiverIO backbone. We also adapt \textbf{Spatial Forcing}~\citep{li2026spatial} by unprojecting a frozen \textbf{VGGT-$\Omega$}~\citep{wang2026vggt} representation into calibrated 3D voxels and distilling it into the same PerceiverIO backbone. These two baselines serve as contrasts to our amodal GEM3D representation, allowing us to isolate the benefits of our representation from those of the distillation strategy.
\vspace{-0.2in}
\vspace{0.12in}
\subsection{Experimental Results}
\label{sec:experimental_results}

\subsubsection{Comparison with State-of-the-Art Methods}
\label{sec:comparison_sota}

\textbf{In-domain evaluation.}
In real-world experiments, \Reffig{fig:real_results} shows that \MethodName achieves an average Stage II full-task success rate of 88.3\% across the three tasks, outperforming \textit{Spatial Forcing} (45.0\%) by 43.3 points and \textit{PerAct} (41.7\%) by 46.6 points.
In simulation, \Reftab{tab:main} reports six representative RLBench tasks and the 12-task average, where \MethodName obtains an average success rate of 44.2\%, improving over the strongest baseline, \textit{Spatial Forcing} (35.1\%), by 9.1 points.
\textit{ManiGaussian (w Distill)} improves the original \textit{ManiGaussian} from 31.5\% to 33.3\% but remains 10.9 points below \MethodName, showing that both distillation and geometry-grounded representations matter for manipulation.

The weak real-world performance of \textit{ManiGaussian} in \Reffig{fig:real_results} suggests that joint reconstruction and control learning is unstable under noisy, incomplete physical observations. Its rendering objective can compete with action supervision, and the degraded reconstructions in \Reffig{fig:viz_sim} and \Reffig{fig:manigaussian_real_rendering} support this explanation.

\textbf{Generalization evaluation.}
We further evaluate view generalization on eight RLBench tasks (\Reftab{tab:main_view_full_8_tasks}): \textit{PerAct} drops from 29.7\% at the training view to 9.9\%/5.0\% under moderate/large shifts, while \textit{DP3} drops from 42.2\% to 18.6\%/11.5\%. Although 3D policies are commonly expected to be visually robust, a viewpoint change reshapes the partially observed point cloud even after world-frame calibration, causing severe policy degradation under larger shifts. In comparison, \MethodName reaches 41.0\%/29.5\%, exceeding the strongest baselines by 11.7/9.4 points.

Since real-world evaluation includes unseen tabletop textures and object colors, the high training-view success also reflects appearance robustness inherited from diverse RH20T pretraining. Under shifted cameras, \MethodName achieves 78.3\% Stage II success for moderate shifts and 63.3\% for large shifts, compared with 36.7\%/15.0\% for \textit{PerAct} and 30.0\%/6.7\% for \textit{Spatial Forcing}.
Our adapted \textit{Spatial Forcing} adds VGGT-$\Omega$ distillation to the PerAct backbone and improves training-view success from 41.7\% to 45.0\%, yet lowers OOD success below PerAct. We hypothesize that VGGT-$\Omega$ reconstructs only content visible from the current camera, so distilling its view-specific incomplete geometry biases the policy toward the training-view observation pattern. In contrast, \MethodName targets amodal completion and encourages consistent full-scene structure inference, yielding representations less sensitive to input appearance and viewpoint changes. On the first two tasks, its Stage II success across the three views has standard deviations of only 10.8 points for \textit{Place Phone} and 8.5 points for \textit{Stack Cups}.

\textbf{Reconstruction Analysis.}
\Reffig{fig:viz_sim} and \Reffig{fig:viz_real} show that \PretrainingName recovers clearer boundaries and denser geometry than \textit{ManiGaussian}; its 22.17--24.45 dB PSNR also substantially exceeds 13.84--17.02 dB, supporting stronger scene reconstruction for manipulation.

\input{tables/ablation}

\subsection{Ablations}
\label{sec:ablation}
\begin{figure}[!t]
\centering
\begin{minipage}[t]{0.38\textwidth}
    \centering
    \includegraphics[height=3.30cm]{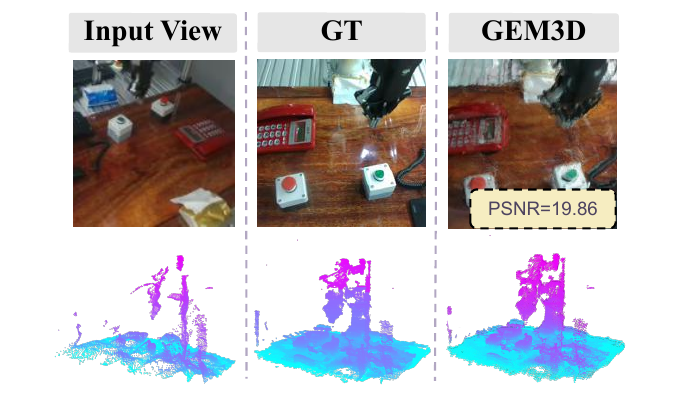}
    \caption{
        \textbf{Image and point cloud reconstruction quality in the real world.}
        }
    \label{fig:viz_real}
\end{minipage}
\hfill
\begin{minipage}[t]{0.6\textwidth}
    \centering
    \includegraphics[height=3.41cm]{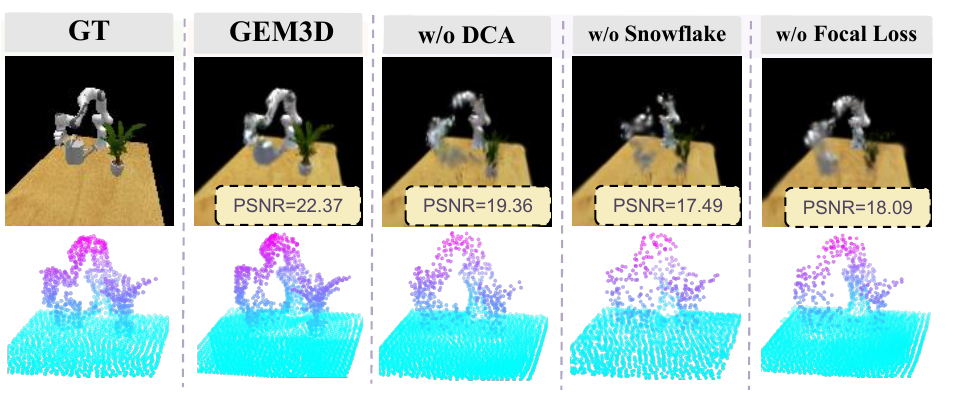}
    \caption{
        \textbf{Qualitative ablation of \PretrainingName designs.}
        Novel-view rendering and point cloud reconstruction results for ablated variants.}
    \label{fig:ablation_rendering}
\end{minipage}
\end{figure}

\textbf{Ablation on GEM3D Pretraining}
We first ablate the architectural designs in \PretrainingName through three variants in \Reffig{fig:ablation_rendering}: \textbf{\textit{w/o DCA}}, which replaces deformable cross-attention with standard cross-attention between dense and downsampled volumetric features; \textbf{\textit{w/o Snowflake}}, which replaces coarse-to-fine point cloud reconstruction with single-step reconstruction; and \textbf{\textit{w/o Focal Loss}}, which replaces focal loss with MSE for Gaussian rendering.
Removing any component clearly degrades reconstruction quality, causing collapsed robot arms, missing plant structures, and less consistent textures.

We further study the dependence on multi-view supervision by varying the number of pretraining views and conducting multi-task training on \texttt{Phone on Base}, \texttt{Meat on Grill}, and \texttt{Turn Tap}.
As shown in \Reftab{tab:ablation_view_nums}, reducing the number of views from eight to two causes only marginal performance drops across viewpoint conditions, suggesting that \PretrainingName remains effective under limited multi-view supervision.
We also include a one-view setting, where Gaussian Splatting rendering is disabled because no held-out view is available for novel-view supervision.
This setting leads to a nearly 10\% drop, showing that Gaussian Splatting is critical for capturing fine-grained appearance details and enforcing cross-view spatial consistency.

\textbf{Ablation on GEM3D Policy}
We ablate the \PolicyName architecture by comparing two variants: \textit{Pretrain-Finetune}, which directly fine-tunes the pretrained encoder without distillation, and \textit{Single-step Distillation}, which removes the $t{+}1$ latent distillation step and aligns only input-level tokens.

\Reftab{tab:ablation_policy} reports the average success rates across three tasks.
The results indicate that incorporating distillation at the $t{+}1$ step improves temporal consistency in the latent dynamics, leading to noticeably stronger policy performance.
In contrast, conventional pretrain-finetune suffers from representation drift due to the domain gap between pretraining and downstream tasks~\citep{hansen2022pre,pang2023masked}, substantially reducing manipulation success.
\clearpage
\endgroup

%% file: tables/sim_results.tex
\begin{table}[!t]
\centering
\small
\captionsetup{skip=4pt}
\caption{
    \textbf{Simulation experimental results.}
    Comparison of \MethodName with state-of-the-art methods using the training view for in-domain evaluation (SR\%).
    Avg. SR is computed over all 12 RLBench tasks, and the remaining columns show six representative tasks.
    \textbf{Bold} numbers denote the best performance, and \underline{underlined} numbers denote the second best.
}
\label{tab:main}
\setlength{\tabcolsep}{4pt}
\renewcommand{\arraystretch}{0.65}
\newcommand{\simresultheaderlinesep}{0.15ex}
\newcommand{\simresultafterheader}{-0.45ex}
\newcommand{\simresultheadone}[1]{#1}
\newcommand{\simresultheadtwo}[2]{\shortstack[c]{#1\\[\simresultheaderlinesep]#2}}
\resizebox{\textwidth}{!}{%
\begin{tabular}{@{}>{\centering\arraybackslash}m{8.8em}*{7}{>{\centering\arraybackslash}m{5em}}@{}}
\toprule
\simresultheadone{\textbf{Methods}} &
\simresultheadtwo{\textbf{12-task}}{\textbf{Avg. SR}} &
\simresultheadtwo{\texttt{Take Money}}{\texttt{out Safe}} &
\simresultheadtwo{\texttt{Close}}{\texttt{Jar}} &
\simresultheadtwo{\texttt{Meat on}}{\texttt{Grill}} &
\simresultheadtwo{\texttt{Water}}{\texttt{Plants}} &
\simresultheadtwo{\texttt{Turn}}{\texttt{Tap}} &
\simresultheadtwo{\texttt{Put Money}}{\texttt{in Safe}} \\[\simresultafterheader]
\midrule
\textit{PerAct} & 21.9 \scriptsize$\pm$ 5.3 & 13.7 \scriptsize$\pm$ 2.9 & 19.0 \scriptsize$\pm$ 4.6 & 16.0 \scriptsize$\pm$ 7.9 & 8.3 \scriptsize$\pm$ 2.3 & 43.7 \scriptsize$\pm$ 4.7 & 10.7 \scriptsize$\pm$ 0.6 \\
\textit{GNFactor} & 29.2 \scriptsize$\pm$ 18.4 & 19.0 \scriptsize$\pm$ 4.0 & \underline{32.0 \scriptsize$\pm$ 16.8} & 35.0 \scriptsize$\pm$ 6.1 & \underline{22.0 \scriptsize$\pm$ 4.4} & 40.7 \scriptsize$\pm$ 10.7 & 11.0 \scriptsize$\pm$ 10.5 \\
\textit{ManiGaussian} & 31.5 \scriptsize$\pm$ 8.5 & \underline{49.3 \scriptsize$\pm$ 10.5} & 27.3 \scriptsize$\pm$ 16.4 & 30.7 \scriptsize$\pm$ 12.9 & 19.3 \scriptsize$\pm$ 9.0 & 44.0 \scriptsize$\pm$ 10.5 & 5.7 \scriptsize$\pm$ 3.5 \\
\textit{DP3} & 33.0 \scriptsize$\pm$ 7.6 & 31.6 \scriptsize$\pm$ 7.1 & 29.7 \scriptsize$\pm$ 8.6 & \underline{55.5 \scriptsize$\pm$ 8.4} & 21.6 \scriptsize$\pm$ 6.4 & \underline{46.8 \scriptsize$\pm$ 8.7} & 15.4 \scriptsize$\pm$ 4.6 \\
{\scriptsize\textit{ManiGaussian (w Distill)}} & 33.3 \scriptsize$\pm$ 10.4 & 43.0 \scriptsize$\pm$ 23.4 & 31.0 \scriptsize$\pm$ 12.1 & 48.7 \scriptsize$\pm$ 2.9 & 16.3 \scriptsize$\pm$ 8.7 & 42.7 \scriptsize$\pm$ 14.6 & 20.3 \scriptsize$\pm$ 9.2 \\
\textit{Spatial Forcing} & \underline{35.1 \scriptsize$\pm$ 7.9} & 46.9 \scriptsize$\pm$ 10.1 & 31.4 \scriptsize$\pm$ 9.4 & 47.1 \scriptsize$\pm$ 8.1 & 21.3 \scriptsize$\pm$ 6.3 & 46.5 \scriptsize$\pm$ 9.0 & \underline{24.1 \scriptsize$\pm$ 7.3} \\
\textit{\textbf{GEM3D} \scriptsize{(Ours)}} & \textbf{44.2 \scriptsize$\pm$ 6.2} & \textbf{72.0 \scriptsize$\pm$ 13.7} & \textbf{44.7 \scriptsize$\pm$ 3.2} & \textbf{61.0 \scriptsize$\pm$ 22.1} & \textbf{32.7 \scriptsize$\pm$ 3.5} & \textbf{54.0 \scriptsize$\pm$ 7.2} & \textbf{29.7 \scriptsize$\pm$ 2.9} \\
\bottomrule
\end{tabular}%
}
\end{table}

%% file: tables/ablation.tex
\begin{table}[!t]
\centering
\caption{\textbf{Ablations on \PretrainingName and \PolicyName.} We select three representative RLBench tasks and report the average success rate under different viewpoint shifts (SR\%).}
\label{tab:ablation}
\newcommand{\ablationsubtablewidth}{0.49\textwidth}
\newcommand{\ablationtabcolsep}{2.2pt}
\newlength{\ablationheaderheight}
\newlength{\ablationheaderdepth}
\newlength{\ablationbodyheight}
\newlength{\ablationpretrainrowheight}
\newlength{\ablationpretrainrowdepth}
\newlength{\ablationpolicyrowheight}
\newlength{\ablationpolicyrowdepth}
\setlength{\ablationheaderheight}{3.2ex}
\setlength{\ablationheaderdepth}{1.4ex}
\setlength{\ablationbodyheight}{14.2ex}
\setlength{\ablationpretrainrowheight}{\dimexpr\ablationbodyheight/4\relax}
\setlength{\ablationpretrainrowdepth}{\dimexpr\ablationpretrainrowheight/3\relax}
\setlength{\ablationpolicyrowheight}{\dimexpr\ablationbodyheight/3\relax}
\setlength{\ablationpolicyrowdepth}{\dimexpr\ablationpolicyrowheight/3\relax}
\newcommand{\ablationheaderstrut}{\rule[-\ablationheaderdepth]{0pt}{\ablationheaderheight}}
\newcommand{\ablationpretrainrowstrut}{\rule[-\ablationpretrainrowdepth]{0pt}{\ablationpretrainrowheight}}
\newcommand{\ablationpolicyrowstrut}{\rule[-\ablationpolicyrowdepth]{0pt}{\ablationpolicyrowheight}}

\begin{subtable}[t]{\ablationsubtablewidth}
\centering
\small
\setlength{\tabcolsep}{\ablationtabcolsep}
\begin{tabular*}{\linewidth}{@{\extracolsep{\fill}}c c c c@{}}
\toprule
\ablationheaderstrut\textbf{Num. Views} & Training& Moderate& Large\\
\midrule
\ablationpretrainrowstrut 8 & 58.8 \scriptsize$\pm$ 6.0 & 45.9 \scriptsize$\pm$ 8.4 & 39.6 \scriptsize$\pm$ 8.5 \\
\ablationpretrainrowstrut 4 & 54.6 \scriptsize$\pm$ 4.9 & 42.4 \scriptsize$\pm$ 6.6 & 37.3 \scriptsize$\pm$ 8.8 \\
\ablationpretrainrowstrut 2 & 54.9 \scriptsize$\pm$ 5.1 & 40.3 \scriptsize$\pm$ 5.1 & 33.6 \scriptsize$\pm$ 6.8 \\
\ablationpretrainrowstrut 1 & 47.2 \scriptsize$\pm$ 7.4 & 33.4 \scriptsize$\pm$ 6.1 & 30.8 \scriptsize$\pm$ 5.8 \\
\bottomrule
\end{tabular*}
\caption{Ablation on the number of pretraining viewpoints.}
\label{tab:ablation_view_nums}
\end{subtable}
\hfill
\begin{subtable}[t]{\ablationsubtablewidth}
\centering
\small
\setlength{\tabcolsep}{\ablationtabcolsep}
\begin{tabular*}{\linewidth}{@{\extracolsep{\fill}}c c c c@{}}
\toprule
\ablationheaderstrut\textbf{Methods} & Training& Moderate& Large  \\
\midrule
\ablationpolicyrowstrut\textit{\makecell{Pretrain-\\Finetune}} & 32.2 \scriptsize$\pm$ 11.1 & 21.1 \scriptsize$\pm$ 6.7 & 15.1 \scriptsize$\pm$ 6.8 \\
\ablationpolicyrowstrut\textit{\makecell{Single-step\\Distillation}} & 54.6 \scriptsize$\pm$ 7.6 & 41.7 \scriptsize$\pm$ 9.0 & 37.7 \scriptsize$\pm$ 8.7 \\
\ablationpolicyrowstrut\textit{\makecell{\textbf{GEM3D}\\[-4pt]{\scriptsize(ours)}}} & 58.8 \scriptsize$\pm$ 6.0 & 45.9 \scriptsize$\pm$ 8.4 & 39.6 \scriptsize$\pm$ 8.5 \\
\bottomrule
\end{tabular*}
\caption{Ablation on different policy training paradigms.}
\label{tab:ablation_policy}
\end{subtable}
\end{table}

%% file: sec/5_discussion.tex
\section{Discussion}
\label{sec:discussion}
\vspace{-0.65\baselineskip}
\begin{figure}[H]
    \centering
    \captionsetup[subfigure]{font=scriptsize,skip=1pt}
    \vspace*{6pt}
    \begin{subfigure}[b]{0.718\linewidth}
        \centering
        \hspace*{-1mm}%
        \raisebox{0.95cm}[\height][0pt]{\includegraphics[width=0.405\linewidth]{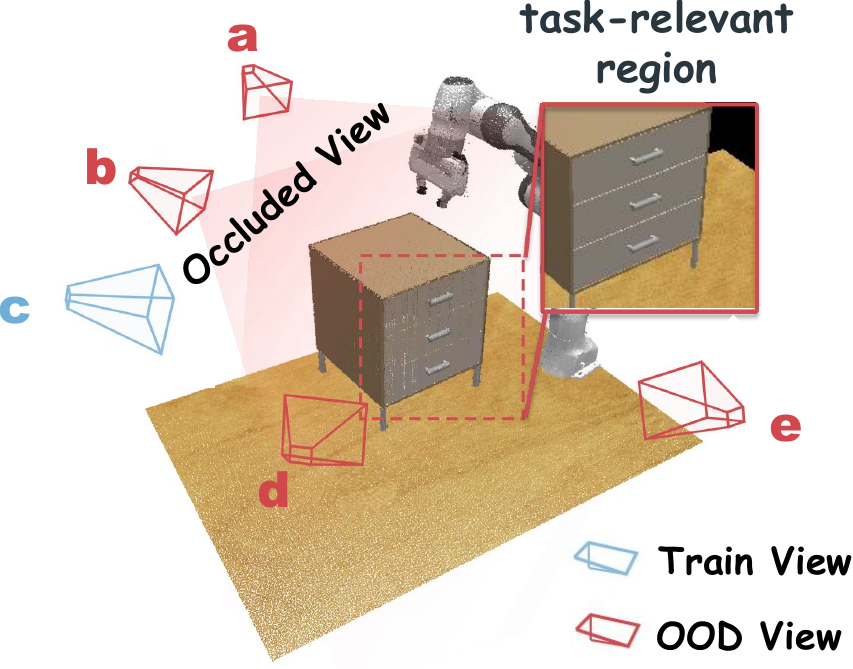}}\hspace*{1.5mm}\hfill%
        \includegraphics[width=0.585\linewidth]{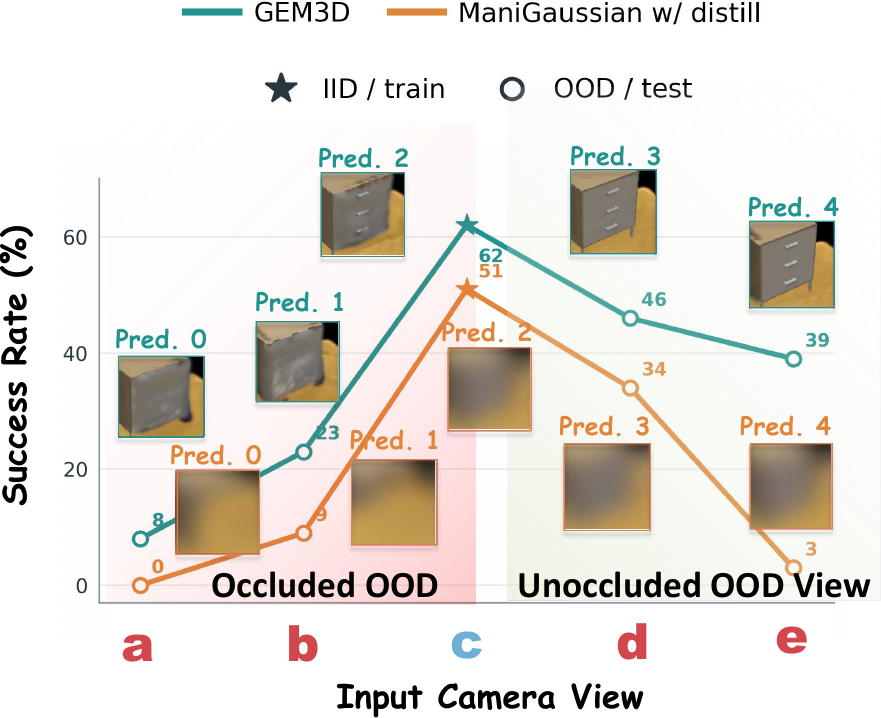}
        \caption{Rendering and policy performance.}
        \label{fig:discussion_rendering}
    \end{subfigure}\hfill%
    \begin{subfigure}[b]{0.266\linewidth}
        \centering
        \includegraphics[width=\linewidth]{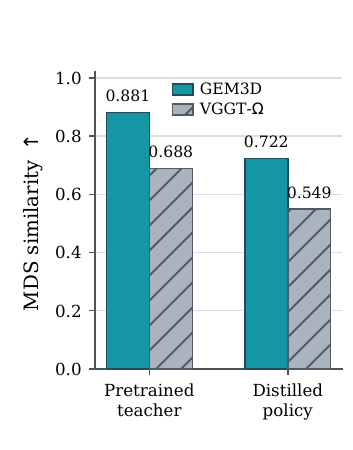}
        \caption{Single-to-multi-view MDS similarity.}
        \label{fig:discussion_view_alignment}
    \end{subfigure}
    \caption{
        \textbf{Rendering, policy performance, and cross-view feature alignment.}
        \textbf{(a)} \MethodNameNs\ and \textit{ManiGaussian} reconstructions and seed-0 success over 100 \textit{Open Drawer} episodes per view.
        \textbf{(b)} Single-to-multi-view MDS similarity on paired \textit{Open Drawer} scene states before and after policy distillation. Higher is better.}
    \label{fig:discussion_analysis}
\end{figure}

\textbf{Q: Do \MethodNameNs's view-generalization gains arise from scene memorization or prior exposure to multiple OOD viewpoints during pretraining?}
No. Pretraining uses 48 RH20T tasks, whereas deployment involves three unseen tasks in our setup (\Reffig{fig:real_results}), with substantially different layouts, equipment, and lighting.
These scenes and their OOD observations are absent from pretraining, ruling out direct memorization.
Our key idea is to learn amodal representations of complete scenes through large-scale reconstruction from partial single-view inputs under multi-view supervision.
Distillation transfers spatial inference and cross-view consistency capabilities to policies trained at only one fixed view, enabling task success, even in novel scenes and from unseen viewpoints.

\textbf{Q: Does improved rendering quality directly imply higher policy success?}
Reconstruction quality in task-critical regions is strongly, though not strictly, related to control performance.
In \Reffig{fig:discussion_analysis}(a), \MethodNameNs\ preserves the ``drawer'' object under OOD views \textbf{d} and \textbf{e}, achieving 46\% and 39\% success; \textit{ManiGaussian} loses this structure and reaches only 34\% and 3\%.
Heavy occlusion in rear views \textbf{a} and \textbf{b} degrades both reconstructions and policies: \MethodNameNs\ reduces viewpoint-induced observation gaps but cannot compensate for missing visual evidence.
Full-scene reconstruction provides task-agnostic supervision for scalable pretraining.

\textbf{Q: Does \MethodNameNs\ learn more view-consistent representations than VGGT-$\Omega$?}
For paired \textit{Open Drawer} states, we jointly project single-view and calibrated multi-view encoder features with MDS into $X,Y\in\mathbb{R}^{N\times 2}$ and compute $S_{\mathrm{MDS}}(X,Y)=\left\|\left(X/\|X\|_F\right)^{\top}\left(Y/\|Y\|_F\right)\right\|_{*}$ (higher is better).
In \Reffig{fig:discussion_analysis}(b), \MethodNameNs\ exceeds VGGT-$\Omega$ before distillation (0.881 vs.\ 0.688) and afterward (0.722 vs.\ 0.549).
Higher single-to-multi-view similarity means that the representation inferred from one partial view is closer to that obtained from the more complete multi-view observation; \MethodNameNs's higher scores therefore indicate stronger global scene understanding and cross-view consistency, helping explain its improved OOD-view control.

%% file: sec/5_conclusion.tex
\section{Conclusion \& Limitations}
\label{sec:conclusion}

We introduce \textit{GEM3D}, a unified representation-policy learning framework for view-generalizable robotic manipulation.
By combining geometrically-grounded amodal 3D pretraining with multi-step policy distillation, \MethodName improves both in-domain manipulation performance and zero-shot view generalization in simulation and real-world experiments.


\MethodName has two main limitations: pretraining relies on potentially costly multi-view RGB-D supervision, and both stages require observations calibrated to a shared world frame, reducing scalability and introducing sensitivity to calibration errors and setup overhead.

%% file: sec/X_suppl.tex
\clearpage
\appendix
\section*{Appendix}
\setcounter{section}{0}
\setcounter{figure}{0}
\setcounter{table}{0}
\setcounter{equation}{0}

\renewcommand{\thesection}{\Alph{section}}
\renewcommand{\thesubsection}{\Alph{section}.\arabic{subsection}}
\renewcommand{\thefigure}{\Alph{section}.\arabic{figure}}
\renewcommand{\thetable}{\Alph{section}.\arabic{table}}
\renewcommand{\theequation}{\Alph{section}.\arabic{equation}}
\section{Method Details}
\label{sec:method_details}
This section provides implementation details for \PretrainingName, complementing the pipeline overview in \Reffig{fig:pretrain2}.

\subsection{Objective Details}
\label{sec:objectives_appendix}
The implementation stores the voxel feature as $F_t\in\mathbb{R}^{B\times C\times D\times D\times D}$ and flattens its spatial axes to the $D^3\times C$ notation used in the main paper. All target depth maps are back-projected to the shared world frame and cropped to the task workspace before constructing $P_{\mathrm{full}}$.

\paragraph{Multi-scale reconstruction.}
Let $n_i=|\hat P_i|$ and $n_{\max}=\max_i n_i$. We sample $\bar P=\operatorname{FPS}(P_{\mathrm{full}},n_{\max})$ once and use $P_i=\bar P_{1:n_i}$ as the target at scale $i$. The implementation uses the symmetric Chamfer-
\begin{equation}
\mathrm{CD}(A,B)=\frac{1}{2}\left(
\frac{1}{|A|}\sum_{a\in A}\min_{b\in B}\|a-b\|_2+
\frac{1}{|B|}\sum_{b\in B}\min_{a\in A}\|b-a\|_2
\right),
\end{equation}
with scale weights $(w_0,w_1,w_2,w_3)=(2.5,1.5,1.25,2.0)$. The distances are not squared.

\paragraph{Gaussian rendering.}
For anchor $\hat p_j$, the Gaussian head predicts center, opacity, anisotropic scale, rotation, and spherical-harmonic color as
\begin{equation}
\mu_j=\hat p_j+\eta\Delta\mu_j,\quad
\alpha_j=\operatorname{sigmoid}(o_j),\quad
s_j=\exp(\ell_j),\quad
q_j=r_j/\|r_j\|_2.
\end{equation}
For an image pair $(X,Y)$, the focal-reweighted MSE is
\begin{equation}
\ell_{\mathrm{fMSE}}(X,Y)=\operatorname{mean}\!\left[\left(1-e^{-|X-Y|}\right)^2(X-Y)^2\right].
\end{equation}
For each valid target view, it is applied to both the full RGB image and the depth-valid region $M_v=\mathbf{1}[D_v>0]$. Masked LPIPS and DSSIM terms are added after rendering starts; padded views are excluded from all reductions. During the geometry-only warm-up, $\mathcal L_{\mathrm{rend}}=0$.

\paragraph{Policy distillation and action loss.}
The frozen teacher is evaluated without gradients. A $5^3$ patchification with stride $5$ converts each $100^3$ voxel feature into 8,000 patch tokens, while adaptive max pooling produces one global token. Let $(p_t,g_t)$ and $(\bar p_{t+1},\bar g_{t+1})$ be the current and predicted-next policy tokens, and $(\tilde p_t,\tilde g_t)$ the corresponding teacher tokens. With $\mathcal L_{\mathrm{cos}}(x,y)=1-\operatorname{mean}(\operatorname{cosine}(x,y))$, the expanded multi-step objective is
\begin{equation}
\begin{aligned}
\mathcal L_{\mathrm{multi\text{-}distill}}={}&
\mathcal L_{\mathrm{cos}}(p_t,\operatorname{sg}[\tilde p_t])+
\mathcal L_{\mathrm{cos}}(g_t,\operatorname{sg}[\tilde g_t])\\
&+\mathcal L_{\mathrm{cos}}(\bar p_{t+1},\operatorname{sg}[\tilde p_{t+1}])
+\mathcal L_{\mathrm{cos}}(\bar g_{t+1},\operatorname{sg}[\tilde g_{t+1}]).
\end{aligned}
\end{equation}
The action loss is the sum of cross-entropy losses for translation, three discretized rotation axes, gripper state, and collision state:
\begin{equation}
\mathcal L_{\mathrm{action}}=
\mathcal L_{\mathrm{CE}}^{\mathrm{trans}}+
\sum_{k\in\left\lbrace x,y,z\right\rbrace}\mathcal L_{\mathrm{CE}}^{\mathrm{rot},k}+
\mathcal L_{\mathrm{CE}}^{\mathrm{grip}}+
\mathcal L_{\mathrm{CE}}^{\mathrm{collision}}.
\end{equation}
All action terms use unit weight. The distillation coefficient follows the configured decreasing schedule $[1.0,0.5,0.3,0.2,0.1,0.05]$.

\subsection{Implementation Details of \textbf{\textit{3D Feature Extractor}}}
\label{sec:3d_feature_extractor_appendix}
Given a single-view observation $o^t = \{\mathcal{I}^t,\; \mathcal{D}^t,\; \mathcal{K}^t\}$ at timestep $t$, the camera parameters $\mathcal{K}^t$ include a camera-to-world extrinsic matrix
$E_{\mathrm{c2w}} =
\begin{bmatrix}
R & t \\
\mathbf{0}^{\mathsf T} & 1
\end{bmatrix}$
and the intrinsic matrix
$K =
\begin{bmatrix}
f_x & 0 & c_x \\
0 & f_y & c_y \\
0 & 0 & 1
\end{bmatrix}$.
The extrinsic maps homogeneous camera coordinates to the shared world frame. For datasets that store a world-to-camera pose, we invert it before back-projection.

The depth map $\mathcal{D}^t$ is back-projected to 3D in two steps.
First, each pixel $(u, v)$ is lifted into the camera coordinate system using the intrinsics $K$:
\begin{equation}
\label{eq:backproj_cam}
\begin{bmatrix}
x_c \\[2pt]
y_c \\[2pt]
z_c
\end{bmatrix}
=
\begin{bmatrix}
\frac{(u - c_x)\, \mathcal{D}^{t}(u, v)}{f_x} \\[8pt]
\frac{(v - c_y)\, \mathcal{D}^{t}(u, v)}{f_y} \\[8pt]
\mathcal{D}^{t}(u, v)
\end{bmatrix}.
\end{equation}

The resulting point is transformed into the shared world coordinate system in homogeneous coordinates:
\begin{equation}
\label{eq:backproj_world}
\begin{bmatrix}
x \\[2pt]
y \\[2pt]
z \\[2pt]
1
\end{bmatrix}
=
E_{\mathrm{c2w}}
\begin{bmatrix}
x_c \\[2pt]
y_c \\[2pt]
z_c \\[2pt]
1
\end{bmatrix}.
\end{equation}

Point clouds back-projected from depth maps typically contain redundant observations, such as background and floor points. We therefore first \textbf{crop} the raw point cloud and apply \textbf{farthest point sampling (FPS)} to regularize its spatial density and obtain a fixed number of points for batched training. The processed point cloud is denoted as $\mathcal{P}$.

\begin{figure}[h]
    \centering
    \includegraphics[width=0.75\linewidth]{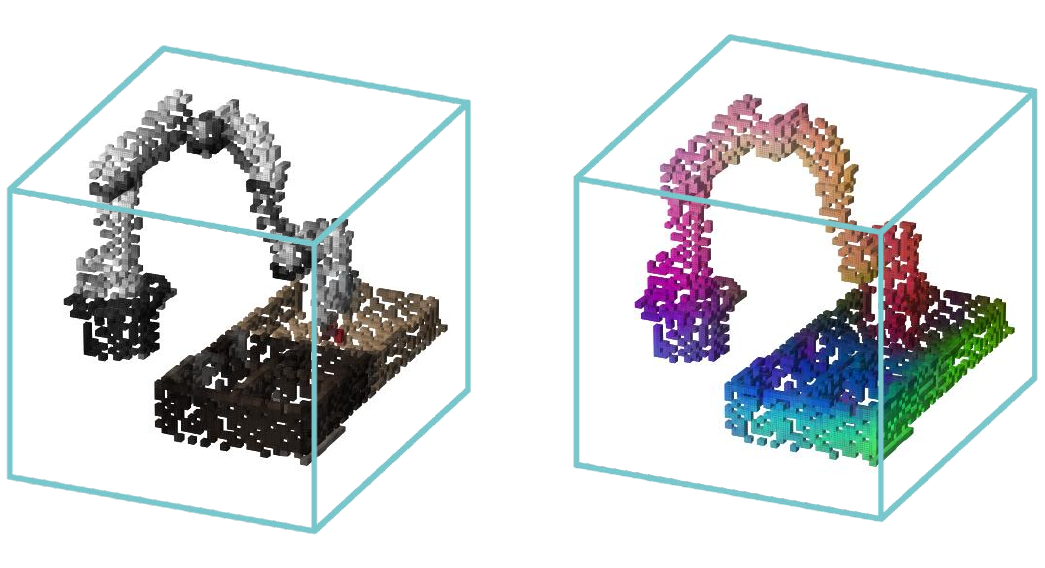}
    \caption{\textbf{Volumetric observation of the single-view input $o^t$}. The left is the \textit{\textbf{3D Occupancy Map}} of size $(d^3, 10)$, and the right is the \textit{\textbf{3D Feature Map}} of size $(d^3, 384)$.}

    \label{fig:voxelgrid_supp}
\end{figure}

However, point clouds are inherently unstructured, which makes global reasoning and occlusion-aware completion difficult. To obtain a structured spatial representation, we introduce a dense volumetric intermediate representation. Specifically, each point in $\mathcal{P}$ is associated with its RGB value and the corresponding DinoV2 feature extracted from the input image. The resulting occupancy and feature-augmented point clouds are jointly voxelized to produce two volumetric observations: a \textit{\textbf{3D Occupancy Map}} and a \textit{\textbf{3D Feature Map}} (see~\Reffig{fig:voxelgrid_supp}).

Feature extraction is performed in the voxelized space using a \textbf{fusion-style 3D U-Net}, which combines the \textit{\textbf{3D Occupancy Map}} and the \textit{\textbf{3D Feature Map}} to produce a dense volumetric feature $\mathcal{F}$ of size $(D^{3},128)$.

The overall procedure is summarized in Algorithm~\ref{alg:voxel_pipeline}.
\begin{algorithm}[!htbp]
\DontPrintSemicolon
\SetAlgoLined
\caption{3D Feature Extractor}
\label{alg:voxel_pipeline}

\KwIn{$\mathcal{I}^t$, $\mathcal{D}^t$, $\mathcal{K}^t=\{E_{\mathrm{c2w}},K\}$}
\KwOut{Dense volumetric feature $\mathcal{F}$}

$\mathcal{F}_{2D} \gets \texttt{DinoV2}(\mathcal{I}^t)$
\hspace{2mm}\textcolor{gray}{// ($H \times W, 384$)}

$P_{\text{raw}} \gets \textsc{Pixel2Point}(\mathcal{D}^t, \mathcal{K}^t)$ \textcolor{gray}{// Eq.~\ref{eq:backproj_cam}, Eq.~\ref{eq:backproj_world}} \;

$P_{\text{crop}} \gets \textsc{Crop}(P_{\text{raw}})$\;
$P \gets \textsc{FPS}(P_{\text{crop}}, N)$\;


$(V_{\text{occ}}, V_{\text{feat}}) \gets \textsc{Voxelize}(P, \mathcal{I}^t, \mathcal{F}_{2D})$
\textcolor{gray}{// $(D^3,10)$, $(D^3,384)$}

$\mathcal{F} \gets \texttt{3D UNet}(V_{\text{occ}}, V_{\text{feat}})$ \textcolor{gray}{// $(D^3,128)$}

\Return $\mathcal{F}$\;

\end{algorithm}

\subsection{Implementation Details of \textit{\textbf{3D Deformable Attention}}}
\label{sec:coarse2fine_appendix}
In our \textbf{coarse-to-fine point cloud reconstruction} module, we adopt a \textbf{3D Deformable Cross-Attention (DCA)} mechanism operating over voxel proposal queries.
Specifically, the input $x$ corresponds to the \emph{proposal voxel queries} $\mathcal{Q}_p \in \mathbb{R}^{N_q \times C}$ from the preceding stage, where $N_q=d^3$ is the number of query voxels and $C$ is the channel dimension (here $C=128$). We denote the number of sampled offsets per query by $K$.
The context features are provided by the dense volumetric feature $\mathcal{F}$ extracted from the fusion-style 3D U-Net.

For each voxel query token $\mathbf{q}_p \in \mathcal{Q}_p$, the module predicts a set of learnable offsets
$\{\Delta \mathbf{p}_k(\mathbf{q}_p)\}_{k=1}^{K}$ around a reference voxel position $\mathbf{p}_0$, forming a \textit{voxel-wise sampling grid} $v_\text{grid}$.
Keys and values are sampled from $\mathcal{F}$ at these locations using \texttt{F.grid\_sample}. The sampled values are then aggregated with attention weights $A_k(\mathbf{q}_p)$ to capture local geometric context.
The refined token feature, referred to as the \emph{seed token}, is obtained as:

\setlength{\abovedisplayskip}{3pt}
\setlength{\belowdisplayskip}{3pt}
\begin{equation}
\text{DCA}(\mathbf{q}_p, \mathcal{F})
= \sum_{k=1}^{K}
\tilde{A}_k(\mathbf{q}_p)\, \mathcal{F}(\mathbf{p}_0 + \Delta \mathbf{p}_k(\mathbf{q}_p)).
\end{equation}

To enhance 3D geometric awareness, a \textbf{continuous positional bias (CPB)} is integrated.
For each query voxel token $\mathbf{q}_p$ and its sampled locations $\mathbf{p}_0 + \Delta \mathbf{p}_k(\mathbf{q}_p)$, CPB produces a learnable bias
\begin{equation}
b_{p,k} = \text{CPB}\big(\mathbf{q}_p, \mathbf{p}_0 + \Delta \mathbf{p}_k(\mathbf{q}_p)\big),
\end{equation}
which is directly added to the raw attention logits $e_k(\mathbf{q}_p)$ before normalization:
\begin{equation}
\tilde{A}_k(\mathbf{q}_p) = \text{softmax}_k\big( e_k(\mathbf{q}_p) + b_{p,k} \big),
\end{equation}

Together with the learned offsets, this design allows DCA to selectively aggregate features from spatially relevant voxels and refine each voxel token with local geometric context.
In essence, CPB provides a learnable bias that modulates attention weights based on relative 3D positions.

Compared to standard full-grid 3D cross-attention, which has complexity $O(N_q D^3 C)$ for $N_q$ queries and $D^3$ voxels, DCA reduces the attention cost to $O(N_q K C)$ by attending to only $K\ll D^3$ sampled locations per query.

\Refalg{alg:dca_voxel} summarizes the implementation procedure of 3D DCA.

\begin{algorithm}[!htbp]
\DontPrintSemicolon
\SetAlgoLined
\caption{3D Deformable Cross-Attention}
\label{alg:dca_voxel}

\KwIn{Voxel proposal queries $\mathcal{Q}_p \in \mathbb{R}^{N_q \times C}$, dense volumetric feature $\mathcal{F}$, number of offsets $K$}
\KwOut{Refined voxel tokens $\mathcal{Q}_p^\text{refined}$}

$\{\Delta \mathbf{p}_k(\mathbf{q}_p)\}_{k=1}^{K} \gets \textsc{OffsetPredictor}(\mathcal{Q}_p)$
\textcolor{gray}{// Learnable offsets per voxel token}

$\mathbf{p}_k \gets \mathbf{p}_0 + \Delta \mathbf{p}_k(\mathbf{q}_p)$
\textcolor{gray}{// Sampling locations in 3D voxel space}

$\mathsf{K}, \mathsf{V} \gets \texttt{F.grid\_sample}(\mathcal{F}, \mathbf{p}_k)$
\textcolor{gray}{// Extract voxel features at sampled offsets}

$\tilde{A}_k(\mathbf{q}_p) \gets \textsc{Softmax}(\langle \mathbf{q}_p, \mathsf{K} \rangle + \textsc{CPB}(\mathbf{q}_p, \mathbf{p}_k))$
\textcolor{gray}{// Continuous positional bias modulates attention}

$\mathcal{Q}_p^\text{refined} \gets \sum_{k=1}^{K} \tilde{A}_k(\mathbf{q}_p) \, \mathsf{V}_k$
\textcolor{gray}{// Seed tokens capturing local geometry}

\Return $\mathcal{Q}_p^\text{refined}$\;

\end{algorithm}

\textbf{Design Highlights of 3D DCA:}
\begin{itemize}
    \item Adaptive offsets enable attention to focus on locally informative voxels rather than uniform grids.
    \item Dense voxel features are queried via the voxel-wise sampling grid using \texttt{F.grid\_sample}, enabling efficient aggregation of local geometric information from the volumetric feature field to each proposal voxel query.
    \item Continuous positional bias (CPB) encodes relative 3D positions and guides attention toward geometrically relevant locations.
\end{itemize}

\section{Training Details}
\label{sec:training_details_appendix}

\subsection{Training Details for \textbf{\PretrainingName}}
Training on robotic scenes with large appearance, geometry, and motion variation is challenging.
The reconstruction visualizations and ablations in \Reffig{fig:viz_sim}, \Reffig{fig:viz_real}, and \Reffig{fig:ablation_rendering} show that stable geometry and rendering quality are important for downstream policy learning.
In practice, we adopt two training strategies to stabilize feed-forward Gaussian Splatting across diverse scenes:

\begin{itemize}
    \item \textit{Stage-wise decoupled reconstruction and rendering.}
    The quality of Gaussian Splatting reconstruction is highly sensitive to the initialization of point positions. Conventional 3D Gaussian Splatting often relies on point clouds initialized from COLMAP. Similarly, during training, the rendering stage (c) strongly depends on the reconstruction quality of stage (b), especially when both modules share the same feature representation $\mathcal{F}$ to simultaneously solve two related but distinct tasks: geometry reconstruction and novel view synthesis.

    If novel view rendering is enforced too early while the reconstruction quality is still poor, the gradients from stage (c) may negatively interfere with the learning of $\mathcal{F}$, leading to unstable convergence and even complete collapse of the reconstructed scene. Therefore, during the first $\delta$ training iterations, we optimize only the reconstruction objective of stage (b) and disable the rendering module (c). After the model can recover reliable and appearance-consistent point clouds from $\mathcal{F}$, we enable stage (c) and allow gradients from novel-view rendering to be back-propagated. This strategy provides a stable geometric prior for subsequent Gaussian Splatting.

    \item \textit{Multi-step iterative reconstruction per scene.}
    A fundamental difference between feed-forward Gaussian Splatting and traditional optimization-based splatting lies in the need for strong cross-scene generalization. In the early training phase, directly switching to a new scene after a single gradient update often leads to reconstruction failure due to large scene variance.

    To mitigate this issue, we perform $k$ consecutive optimization steps on the same scene before switching to the next one. As training progresses and the model becomes more stable, we gradually decrease $k$ until it eventually becomes $1$, corresponding to standard single-step updates per scene.
\end{itemize}

\subsubsection{Training Strategy for \textbf{\PolicyName}}
During policy learning, imitation is performed from human demonstrations. Raw trajectories often contain redundant intermediate states, such as gripper approaching motions. We therefore select a sparse set of keyframes that correspond to interaction-critical states.

Given a demonstration trajectory
\[
\mathcal{D} = \{o_t\}_{t=1}^{T},
\]
where $g_t \in \{0,1\}$ denotes the gripper open/close state, a timestep $t$ is selected as a keyframe if
\begin{equation}
t \in \mathcal{K} \;\; \Longleftrightarrow \;\;
(g_t \neq g_{t-1}) \;\vee\; (t = T) \;\vee\; \textsc{Stopped}(o_t),
\end{equation}
where $\textsc{Stopped}(\cdot)$ indicates near-static motion under a small velocity threshold.
Redundant adjacent keyframes are further removed. This strategy shortens the demonstration sequence while preserving states that are critical for object interaction.

\subsection{Parameter Details}

\input{tables/pretraining_hyperparameters}

\input{tables/policy_hyperparameters}

\section{Task Descriptions}
\label{sec:tasks_description}

\begin{figure}[h]
    \includegraphics[width=1.0\linewidth]{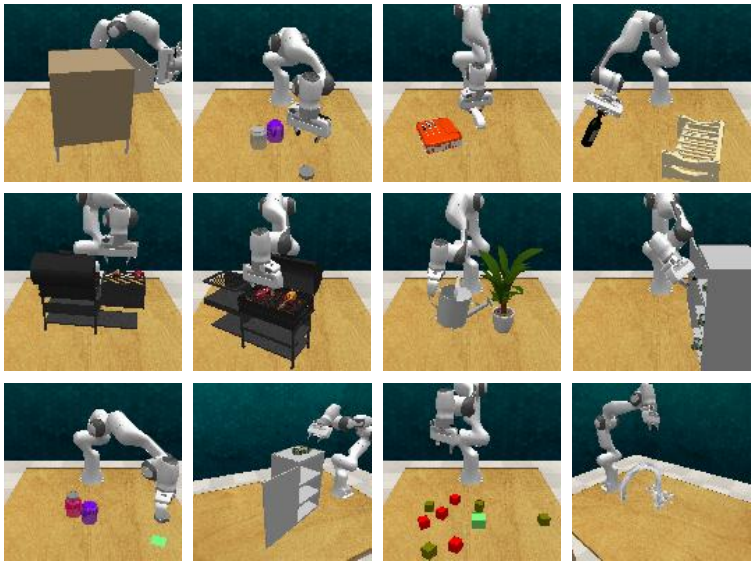}
\caption{\textbf{RLBench tasks}. 12 tasks from nine scenes are selected.}
    \label{fig:12_RLBench_tasks}
\end{figure}

We selected 12 tasks from RLBench, which together contain 178 variations. \Reffig{fig:12_RLBench_tasks} and \Reftab{tab:tasks_description} provide detailed descriptions.

Notably, among these 12 tasks, \texttt{close jar} and \texttt{open jar}, \texttt{meat off grill} and \texttt{meat on grill}, and \texttt{put money in safe} and \texttt{take money out safe} share the same scene but require different action trajectories. Thus, the 12 tasks cover nine distinct scenes. During pretraining, we collect data from nine tasks to cover these scenes, while reserving several action trajectories for downstream policy learning. This design evaluates whether \MethodName can reuse pretrained scene-level geometry for new manipulation trajectories within related environments.

Specifically, the trajectories for \texttt{open jar}, \texttt{meat on grill}, and \texttt{take money out safe} were not seen during pretraining. Nevertheless, the in-domain results in \Reftab{tab:rlbench_full_appendix} show that \MethodName still improves performance on these extrapolated tasks. The view-generalization results in \Reftab{tab:main_view_full_8_tasks} further show that our method remains more robust when the test camera changes, even for tasks with different trajectories within the same scene. This indicates that single-view expert demonstrations during policy learning are sufficient to adapt the pretrained representation to new skills, while preserving scene-level geometric understanding.

\input{tables/tasks_description}

\section{Supplementary Experiments}
\label{sec:supplementary_experiments}

\subsection{Full In-domain RLBench Results}
\label{sec:full_rlbench_results}
\Reftab{tab:rlbench_full_appendix} reports the complete in-domain results on all 12 RLBench tasks. For compactness, the main text shows six representative task columns while computing the average success rate over all 12 tasks. Across the full benchmark, \MethodName achieves the best average SR (\textbf{44.2\%}) and the best average rank, improving over the strongest baseline, \textit{Spatial Forcing} (35.1\%), by \textbf{9.1} points. The gains are especially clear on extrapolated trajectories such as \texttt{meat on grill} and \texttt{take money out safe}, where the task shares the scene with pretraining data but requires a different action sequence.

\input{tables/rlbench_full_results}

\subsection{Inference Efficiency on RLBench}
\label{sec:inference_efficiency_appendix}
\Reftab{tab:inference_speed} reports policy inference speed on the RLBench \texttt{open drawer} task using a single NVIDIA RTX 3090: \MethodName achieves 1.28~Hz, compared with 1.18~Hz for \textit{PerAct} and 1.36~Hz for \textit{ManiGaussian}. At inference, the policy predicts actions from calibrated single-view RGB-D observations without explicit point-cloud reconstruction or Gaussian rendering. Distillation adds computation during policy training, while the frozen teacher and reconstruction branches are not used during deployment.

\input{tables/inference_speed}

\subsection{View Generalization and Occlusion-sensitive Tasks}
\label{sec:additional_details_of_manipulation_tasks}
\begin{figure}[h]
    \centering
    \includegraphics[width=0.65\linewidth]{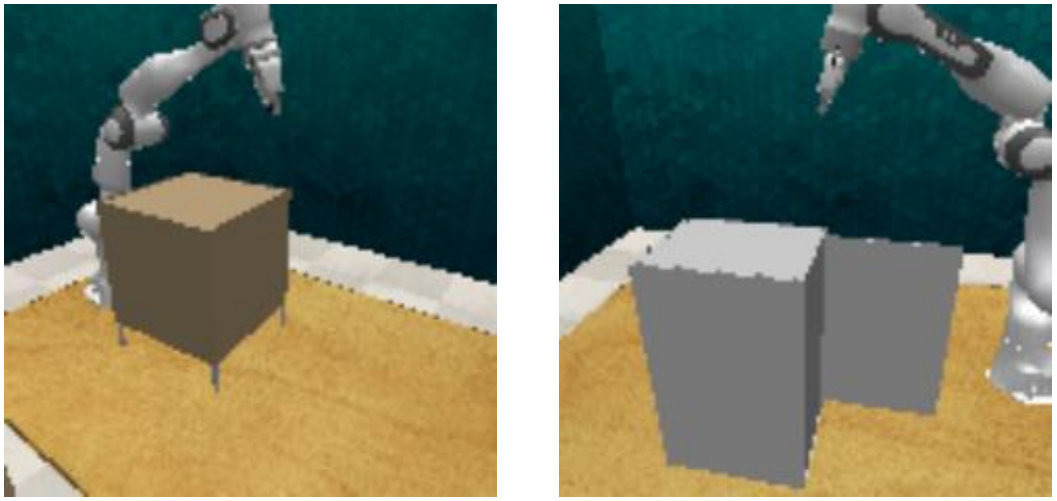}
    \caption{\textbf{\texttt{open drawer} and \texttt{take money out safe} tasks under large view shifts}. Both tasks suffer from severe occlusion, resulting in drastic success rate drops.}
    \label{fig:occ}

\end{figure}

\Reftab{tab:main_view_full_8_tasks} extends the view-generalization evaluation to eight RLBench tasks, including two heavily occluded tasks, \texttt{open drawer} and \texttt{take money out safe}. \MethodName remains the strongest method on average under both moderate and large viewpoint shifts. The performance drop is more pronounced on the two occlusion-sensitive tasks, especially under large shifts, because one side of the scene can become almost fully hidden from the single RGB-D camera, as illustrated in \Reffig{fig:occ}. This indicates that visibility of task-critical contact regions still matters, even when the policy has learned a stronger scene-complete representation.

\input{tables/main_view_full_8_tasks}
\par
\noindent
\begin{minipage}{\textwidth}
    \centering
    \includegraphics[width=0.64\linewidth]{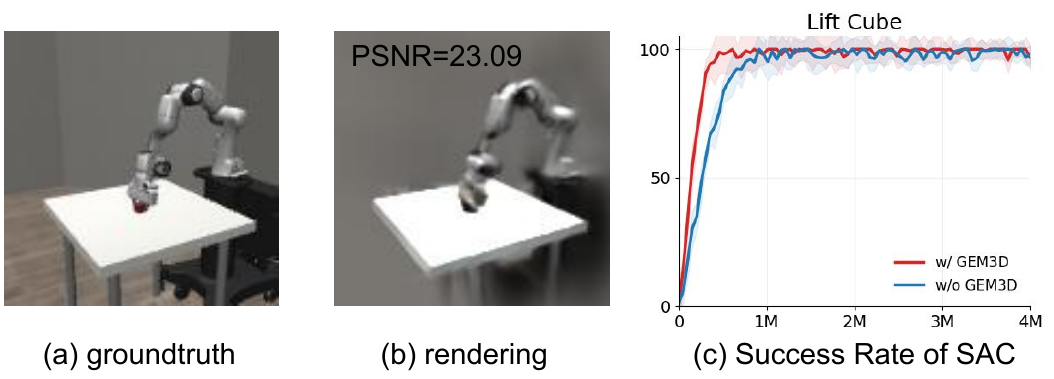}
    \captionof{figure}{\textbf{Robosuite RL transfer.} GEM3D maintains high rendering fidelity in Robosuite and improves downstream RL convergence on the \texttt{Lift} task.}
    \label{fig:robosuite_result}
\end{minipage}
\par

\subsection{Rendering Quality Analysis}
\label{sec:rendering_quality_appendix}
\Reftab{tab:rendering_quality} provides additional quantitative reconstruction results on three representative RLBench tasks. \MethodName improves the average PSNR from 16.05 to \textbf{23.13} and SSIM from 0.3838 to \textbf{0.8136}, while reducing Chamfer $L_2$ from 0.04511 to \textbf{0.01621}. These results are consistent with the visual comparisons in \Reffig{fig:viz_sim} and \Reffig{fig:viz_real}: \MethodName produces sharper novel views and denser, more complete point clouds than \textit{ManiGaussian}.

\begin{figure*}[t]
    \centering

    \begin{subfigure}{\textwidth}
        \centering
        \includegraphics[width=0.9\textwidth]{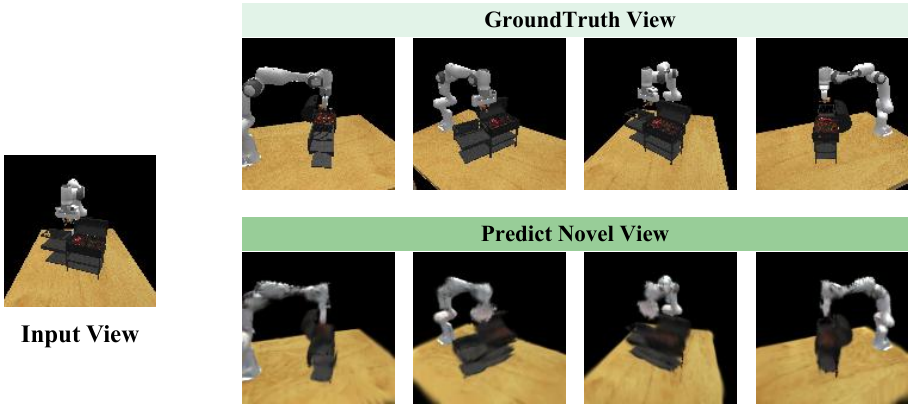}
    \end{subfigure}

    \begin{subfigure}{\textwidth}
        \centering
        \includegraphics[width=0.9\textwidth]{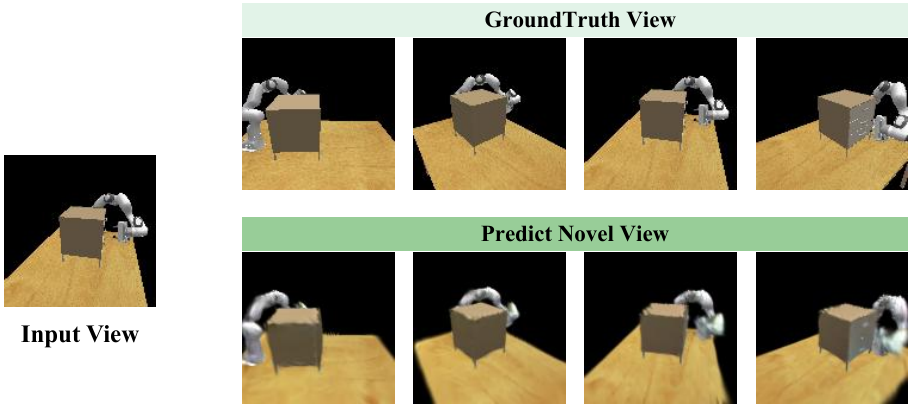}
    \end{subfigure}

    \begin{subfigure}{\textwidth}
        \centering
        \includegraphics[width=0.9\textwidth]{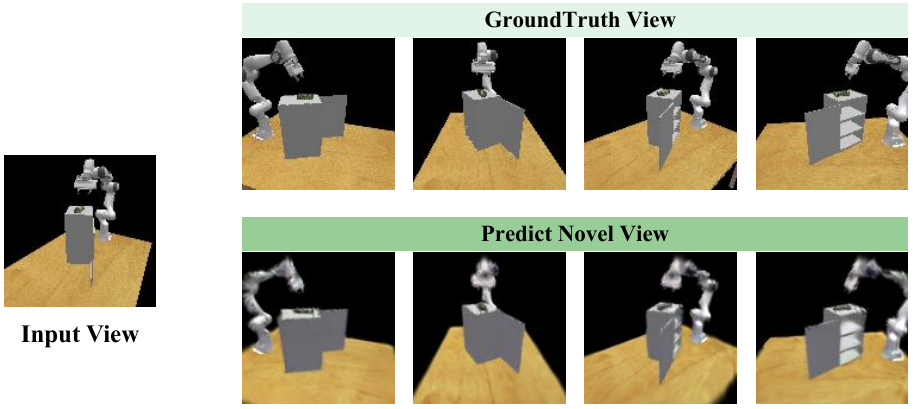}
    \end{subfigure}

    \caption{Additional novel-view rendering results (Part 1/2).}
    \label{fig:vertical_six_a}
\end{figure*}

\begin{figure*}[t]
    \centering

    \begin{subfigure}{\textwidth}
        \centering
        \includegraphics[width=0.9\textwidth]{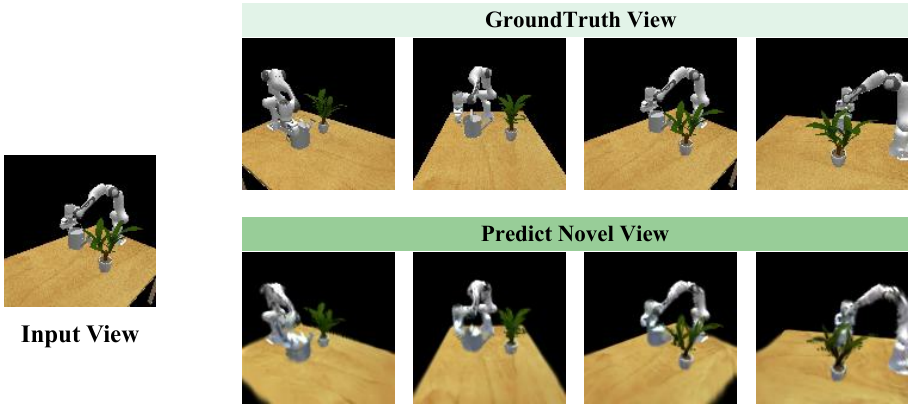}
    \end{subfigure}

    \begin{subfigure}{\textwidth}
        \centering
        \includegraphics[width=0.9\textwidth]{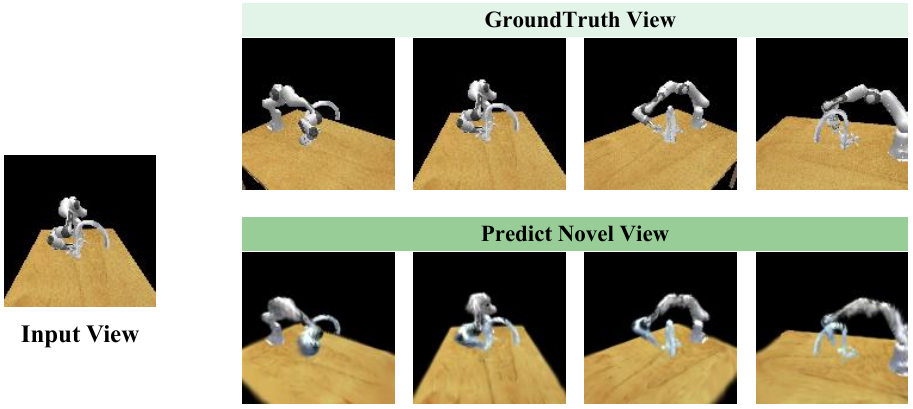}
    \end{subfigure}

    \begin{subfigure}{\textwidth}
        \centering
        \includegraphics[width=0.9\textwidth]{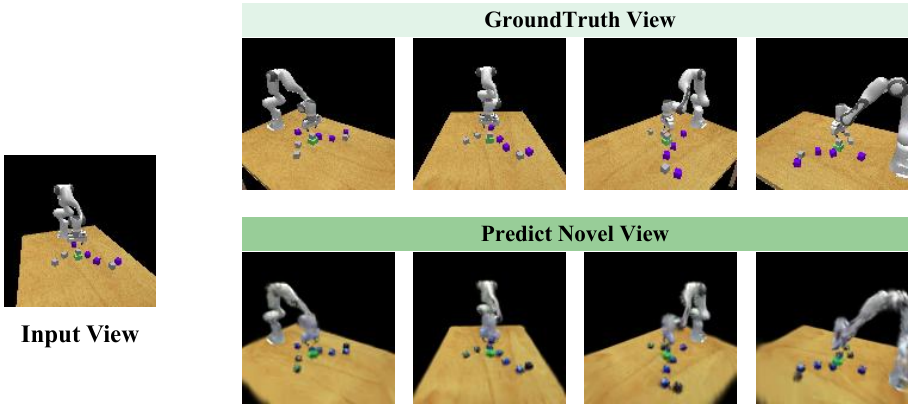}
    \end{subfigure}

    \caption{Additional novel-view rendering results (Part 2/2).}
    \label{fig:vertical_six_b}
\end{figure*}

We further provide qualitative novel-view synthesis results in \Reffig{fig:vertical_six_a} and \Reffig{fig:vertical_six_b}. For each keyframe sequence, novel views are rendered from uniformly distributed virtual cameras and compared across different RLBench tasks.

\input{tables/rendering_quality}

\subsection{Real-world ManiGaussian Rendering}
\label{sec:manigaussian_real_rendering_appendix}
\begin{figure}[H]
    \centering
    \includegraphics[width=0.98\linewidth]{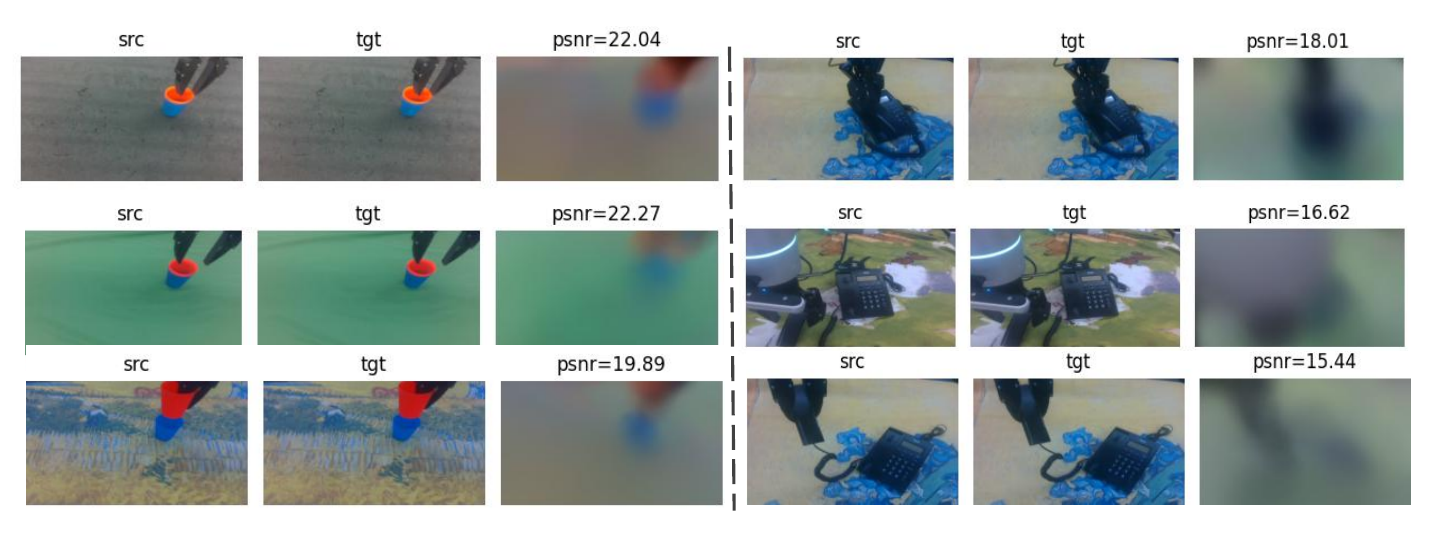}
    \caption{\textbf{Real-world rendering results of \textit{ManiGaussian}.} The source and target views are visually clear, but the rendered novel views are severely blurred and lose task-relevant object geometry.}
    \label{fig:manigaussian_real_rendering}
\end{figure}

\Reffig{fig:manigaussian_real_rendering} visualizes \textit{ManiGaussian} on real-world manipulation scenes.
Although the source and target images are well aligned, the rendered views collapse into low-frequency color blobs: object boundaries are smeared, thin structures such as grippers and phone cables are missing, and contact-relevant foreground objects are often indistinguishable from the tabletop.
This suggests that joint Gaussian rendering and policy learning is difficult under noisy real-world RGB-D observations.
When the reconstruction branch fails to recover reliable geometry and appearance, it can provide weak or misleading supervision for action learning, which helps explain the low real-world success rates of \textit{ManiGaussian} in the main experiments.

\subsection{Multi-task Policy Learning}
\label{sec:multi_task_appendix}
We also evaluate whether the learned representation supports a unified language-conditioned policy trained on multiple tasks. Following the multi-task setting used in our earlier experiments, we train a single policy on a mixture of four RLBench tasks: \texttt{open jar}, \texttt{meat on grill}, \texttt{turn tap}, and \texttt{phone on base}. \Reftab{tab:multi_task} reports the averaged multi-task results, and \Reftab{tab:multi_task_result} reports per-task success rates together with relative changes from the corresponding single-task setting.

\input{tables/multi_task}

Compared with \textit{PerAct}, \MethodName maintains substantially higher success rates under all viewpoint settings. Its average drops from the single-task setting are small under moderate and large shifts, showing that the pretrained geometric representation and distillation-based policy learning remain effective when one policy must distinguish multiple language-conditioned skills.

\input{tables/multi_task_result}

\subsection{Robosuite Reinforcement Learning Transfer}
\label{sec:robosuite_rl_appendix}
To evaluate whether the pretrained representation can also benefit reinforcement learning, we conduct an additional experiment on the \texttt{Lift} task in Robosuite~\citep{zhu2020robosuite}. Robosuite is a modular robot learning benchmark built on MuJoCo and provides controlled manipulation environments with configurable objects, cameras, and reward functions. In this experiment, the robot must lift a cube from the tabletop. We first pretrain the GEM3D visual encoder on an offline Robosuite dataset, then initialize a SAC agent with the distilled policy encoder for downstream RL training.

\Reffig{fig:robosuite_result} shows both reconstruction quality and RL learning performance. The rendering results indicate that the pretrained encoder transfers to the Robosuite visual domain and reconstructs task-relevant cube and robot geometry. More importantly, the SAC agent using the GEM3D-initialized encoder achieves faster and more stable success-rate improvement than training without this representation. This suggests that the learned 3D representation is not limited to imitation learning: it can also provide useful visual features for RL, where sparse reward exploration is sensitive to perception quality and representation stability.

\clearpage

%% file: tables/pretraining_hyperparameters.tex
\begin{table}[htbp]
\centering
\caption{Key Hyperparameters of \textbf{\PretrainingName}}

\label{tab:pretraining_hyperparameters}
\begin{tabular}{l c}
\hline
\textbf{Parameters} & \textbf{Value} \\
\hline
\small\texttt{train.batch\_size} & $1 \times 8$ GPUs \\
\small\texttt{train.learning\_rate} & $1.0 \times 10^{-4}$ \\
\small\texttt{train.optimizer} & torch.optim.AdamW \\
\small\texttt{train.steps} & 50K \\
\small\texttt{train.num\_views} & 8 \\
\small\texttt{train.render\_start\_step} & 4K \\
\small\texttt{train.$k$} & [4, 3, 2, 1] \\
\small\texttt{train.demos} & 50 \\
\small\texttt{network.$D$} & 100 \\
\small\texttt{network.$d$} & [7,7,7] \\
\small\texttt{network.fps\_sample\_num} & 6144 \\
\small\texttt{network.scene\_bounds} & [-0.375,-0.5,0.6,1.0,0.5,1.6] \\
\hline
\end{tabular}
\end{table}

%% file: tables/policy_hyperparameters.tex
\begin{table}[htbp]
\centering
\caption{Key Hyperparameters of \textbf{\PolicyName}}

\label{tab:policy_hyperparameters}
\begin{tabular}{l c}
\hline
\textbf{Parameters} & \textbf{Value} \\
\hline
\small\texttt{train.batch\_size} & 1 \\
\small\texttt{train.learning\_rate} & $1.0 \times 10^{-4}$ \\
\small\texttt{train.optimizer} & torch.optim.AdamW \\
\small\texttt{train.single\_task\_steps} & 12K \\
\small\texttt{train.camera} & [front] \\
\small\texttt{train.demos} & 20 \\
\small\texttt{network.scene\_bounds} & [-0.375,-0.5,0.6,1.0,0.5,1.6]  \\
\small\texttt{network.num\_tokens} & 8000 \\
\small\texttt{network.num\_latents} & 2048 \\
\small\texttt{network.$\lambda_distill$} & [1,0.5,0.3,0.2,0.1,0.05] \\
\hline
\end{tabular}
\end{table}

%% file: tables/tasks_description.tex
\begin{table*}[htbp]
    \centering
    \caption{
        \textbf{Task Descriptions.} The 12 highly challenging tasks include a variety of random variations in color and layout, as well as multiple language instruction variations.
    }
    \label{tab:tasks_description}
    \resizebox{\textwidth}{!}{%
    \begin{tabular}{l c c l}
    \toprule
    Task & Type & Variations & Instruction Template \\
    \midrule

    \texttt{Close Jar} & color & 20 & \textit{``close the \_ jar''} \\
    \texttt{Open Drawer} & placement & 3 & \textit{``open the \_ drawer''} \\
    \texttt{Meat off Grill} & size & 2 & \textit{``take the \_ off the grill''} \\
    \texttt{Turn Tap} & category & 2 & \textit{``turn \_ tap''} \\
    \texttt{Water Plants} & placement & 1 & \textit{``water plant''} \\
    \texttt{Phone on Base} & placement & 1 & \textit{``put the phone on the base''} \\
    \texttt{Put Money in Safe} & placement & 3 & \textit{``put the money away in the safe on the \_ shelf''} \\
    \texttt{Place Wine} & placement & 1 & \textit{``stack wine bottle''} \\
    \texttt{Stack Blocks} & color, count & 60 & \textit{``stack \_ \_ blocks''} \\
    \texttt{Open Jar} & color & 80 & \textit{``open the \_ jar''} \\
    \texttt{Meat on Grill} & category & 2 & \textit{``put the \_ on the grill''} \\
    \texttt{Take Money out Safe} & placement & 3 & \textit{``take the money out of the \_ shelf and place it on the table.''} \\
    
    \bottomrule
    \end{tabular}%
    }
\end{table*}

%% file: tables/rlbench_full_results.tex
\begin{table*}[t]
\centering
\small
\caption{
    Full in-domain results on all 12 RLBench tasks using the training view (SR\%). \textbf{Bold} numbers denote the best performance, and \underline{underlined} numbers denote the second best.
}
\label{tab:rlbench_full_appendix}
\resizebox{\textwidth}{!}{%
\begin{tabular}{c ccccccc}
\toprule
\textbf{Methods} & \makecell{\textbf{Avg.} \\ \textbf{SR (\%)} $\uparrow$} & \makecell{\textbf{Avg.} \\ \textbf{Rank} $\downarrow$} & \makecell{\texttt{Close} \\ \texttt{Jar}} & \makecell{\texttt{Open} \\ \texttt{Drawer}} & \makecell{\texttt{Meat off}\\ \texttt{Grill}} & \makecell{\texttt{Turn} \\ \texttt{Tap}} & \makecell{\texttt{Water} \\ \texttt{Plants}} \\
\midrule
\textit{PerAct} & 21.9 \scriptsize$\pm$ 5.3 & 4.58 & 19.0 \scriptsize$\pm$ 4.6 & 38.7 \scriptsize$\pm$ 13.7 & 28.3 \scriptsize$\pm$ 4.7 & 43.7 \scriptsize$\pm$ 4.7 & 8.3 \scriptsize$\pm$ 2.3 \\
\textit{GNFactor} & 29.2 \scriptsize$\pm$ 18.4 & 3.33 & \underline{32.0 \scriptsize$\pm$ 16.8} & 51.7 \scriptsize$\pm$ 13.5 & 32.0 \scriptsize$\pm$ 10.5 & 40.7 \scriptsize$\pm$ 10.7 & \underline{22.0 \scriptsize$\pm$ 4.4} \\
\textit{ManiGaussian} & 31.5 \scriptsize$\pm$ 8.5 & 3.08 & 27.3 \scriptsize$\pm$ 16.4 & \underline{57.3 \scriptsize$\pm$ 1.5} & 30.0 \scriptsize$\pm$ 11.3 & \underline{44.0 \scriptsize$\pm$ 10.5} & 19.3 \scriptsize$\pm$ 9.0 \\
\textit{ManiGaussian \scriptsize{(w Distill)}} & \underline{33.3 \scriptsize$\pm$ 10.4} & \underline{3.00} & 31.0 \scriptsize$\pm$ 12.1 & 54.3 \scriptsize$\pm$ 12.5 & \underline{35.3 \scriptsize$\pm$ 9.8} & 42.7 \scriptsize$\pm$ 14.6 & 16.3 \scriptsize$\pm$ 8.7 \\
\textit{DP3} & 33.0 \scriptsize$\pm$ 7.6 & 4.17 & 29.7 \scriptsize$\pm$ 8.6 & 33.4 \scriptsize$\pm$ 7.3 & 50.2 \scriptsize$\pm$ 9.1 & 46.8 \scriptsize$\pm$ 8.7 & 21.6 \scriptsize$\pm$ 6.4 \\
\textit{Spatial Forcing} & 35.1 \scriptsize$\pm$ 7.9 & 3.17 & 31.4 \scriptsize$\pm$ 9.4 & 55.0 \scriptsize$\pm$ 9.6 & 38.2 \scriptsize$\pm$ 8.4 & 46.5 \scriptsize$\pm$ 9.0 & 21.3 \scriptsize$\pm$ 6.3 \\
\textit{\textbf{GEM3D} \scriptsize{(Ours)}} & \textbf{44.2 \scriptsize$\pm$ 6.2} & \textbf{1.00} & \textbf{44.7 \scriptsize$\pm$ 3.2} & \textbf{65.7 \scriptsize$\pm$ 4.5} & \textbf{44.3 \scriptsize$\pm$ 3.1} & \textbf{54.0 \scriptsize$\pm$ 7.2} & \textbf{32.7 \scriptsize$\pm$ 3.5} \\
\midrule[\heavyrulewidth]
\textbf{Methods} & \makecell{\texttt{Phone on} \\ \texttt{Base}} & \makecell{\texttt{Put Money} \\ \texttt{in Safe}} & \makecell{\texttt{Place} \\ \texttt{Wine}} & \makecell{\texttt{Stack} \\ \texttt{Blocks}} & \makecell{\texttt{Open}\\ \texttt{Jar}} & \makecell{\texttt{Meat on} \\ \texttt{Grill}} & \makecell{\texttt{Take Money} \\ \texttt{out Safe}} \\
\midrule
\textit{PerAct} & 51.3 \scriptsize$\pm$ 11.9 & 10.7 \scriptsize$\pm$ 0.6 & 3.0 \scriptsize$\pm$ 1.0 & 4.0 \scriptsize$\pm$ 0.0 & 26.7 \scriptsize$\pm$ 9.1 & 16.0 \scriptsize$\pm$ 7.9 & 13.7 \scriptsize$\pm$ 2.9 \\
\textit{GNFactor} & 62.0 \scriptsize$\pm$ 5.6 & 11.0 \scriptsize$\pm$ 10.5 & 5.7 \scriptsize$\pm$ 2.1 & 3.7 \scriptsize$\pm$ 1.2 & 35.7 \scriptsize$\pm$ 6.4 & 35.0 \scriptsize$\pm$ 6.1 & 19.0 \scriptsize$\pm$ 4.0 \\
\textit{ManiGaussian} & \underline{66.3 \scriptsize$\pm$ 12.5} & 5.7 \scriptsize$\pm$ 3.5 & \underline{8.0 \scriptsize$\pm$ 2.6} & 2.0 \scriptsize$\pm$ 1.7 & \underline{38.3 \scriptsize$\pm$ 9.0} & 30.7 \scriptsize$\pm$ 12.9 & \underline{49.3 \scriptsize$\pm$ 10.5} \\
\textit{ManiGaussian \scriptsize{(w Distill)}} & 64.3 \scriptsize$\pm$ 16.2 & \underline{20.3 \scriptsize$\pm$ 9.2} & 5.3 \scriptsize$\pm$ 3.5 & \underline{5.0 \scriptsize$\pm$ 2.6} & 33.7 \scriptsize$\pm$ 9.5 & \underline{48.7 \scriptsize$\pm$ 2.9} & 43.0 \scriptsize$\pm$ 23.4 \\
\textit{DP3} & 63.7 \scriptsize$\pm$ 9.8 & 15.4 \scriptsize$\pm$ 4.6 & 12.7 \scriptsize$\pm$ 3.8 & 8.3 \scriptsize$\pm$ 2.7 & 26.8 \scriptsize$\pm$ 6.9 & 55.5 \scriptsize$\pm$ 8.4 & 31.6 \scriptsize$\pm$ 7.1 \\
\textit{Spatial Forcing} & 63.2 \scriptsize$\pm$ 10.2 & 24.1 \scriptsize$\pm$ 7.3 & 7.4 \scriptsize$\pm$ 3.8 & 4.6 \scriptsize$\pm$ 2.9 & 35.6 \scriptsize$\pm$ 8.7 & 47.1 \scriptsize$\pm$ 8.1 & 46.9 \scriptsize$\pm$ 10.1 \\
\textit{\textbf{GEM3D} \scriptsize{(Ours)}} & \textbf{67.3 \scriptsize$\pm$ 1.5} & \textbf{29.7 \scriptsize$\pm$ 2.9} & \textbf{11.7 \scriptsize$\pm$ 4.0} & \textbf{7.7 \scriptsize$\pm$ 2.9} & \textbf{40.0 \scriptsize$\pm$ 5.6} & \textbf{61.0 \scriptsize$\pm$ 22.1} & \textbf{72.0 \scriptsize$\pm$ 13.7} \\
\bottomrule
\end{tabular}%
}
\end{table*}

%% file: tables/inference_speed.tex
\begin{table}[t]
\centering
\small
\caption{
    Policy inference speed and training cost on the RLBench \texttt{open drawer} task.
    All measurements are taken on a single NVIDIA RTX 3090.
}
\label{tab:inference_speed}
\setlength{\tabcolsep}{4pt}
\begin{tabular*}{\columnwidth}{@{\extracolsep{\fill}} c c c}
\toprule
\textbf{Method} & \textbf{Inference Rate} & \textbf{Policy Training Cost} \\
\midrule
\textit{PerAct} & 1.18 Hz & 2--3 h \\
\textit{ManiGaussian} & 1.36 Hz & 4--5 h \\
\textit{GEM3D Policy} & 1.28 Hz & 5--6 h \\
\bottomrule
\end{tabular*}
\end{table}

%% file: tables/main_view_full_8_tasks.tex
\par\noindent
\begin{minipage}{\textwidth}
    
    \centering
    \captionof{table}{
        Results on eight RLBench tasks at the training view and under moderate/large viewpoint shifts (SR\%). Averages use the same eight tasks in all settings. \textbf{Bold} numbers denote the best performance, and
        $\downarrow$ indicates the relative performance drop compared to the training view.
    }
    \label{tab:main_view_full_8_tasks}
    \small 

    \setlength{\tabcolsep}{1.5pt} 
    
    \resizebox{\textwidth}{!}{%
    \begin{tabular}{c c ccccccccc}
    \toprule
    & \textbf{Methods} & \makecell{Close \\ Jar} & \makecell{Meat off \\ Grill} & \makecell{Turn \\ Tap} & \makecell{Phone on \\ Base} & \makecell{Open \\ Jar} & \makecell{Meat on \\ Grill} & \makecell{Open \\ Drawer}& \makecell{Take Money \\ out Safe} & \textbf{Avg.}\\
    \midrule
    
    \multirowcell{6}[-6pt]{Train\\View}
    & \textit{PerAct} & \var{19.0}{4.6}& \var{28.3}{4.7}& \var{43.7}{4.7}& \var{51.3}{11.9} & \var{26.7}{9.1}& \var{16.0}{7.9} & \var{38.7}{13.7}& \var{13.7}{2.9}& \var{29.7}{13.7}\\
    & \textit{ManiGaussian} & \var{27.3}{16.4} & \var{30.0}{11.3} & \var{44.0}{10.5}& \var{66.3}{12.5} & \var{38.3}{9.0} & \var{30.7}{12.9} & \var{57.3}{1.5}& \var{49.3}{10.5}& \var{42.9}{14.0}\\
    & \textit{\makecell{ManiGaussian \\ [-4pt]{\tiny(w Distill)}}} & \var{31.0}{12.1} & \var{35.3}{9.8}& \var{42.7}{14.6} & \var{64.3}{16.2} & \var{33.7}{9.5} & \var{48.7}{2.9} & \var{54.3}{12.5}& \var{43.0}{23.4}& \var{44.1}{11.3}\\ 
    & \textit{DP3} & \var{29.7}{8.6} & \var{50.2}{9.1} & \var{46.8}{8.7} & \var{63.7}{9.8} & \var{26.8}{6.9} & \var{55.5}{8.4} & \var{33.4}{7.3} & \var{31.6}{7.1} & \var{42.2}{8.1} \\
    & \textit{Spatial Forcing} & \var{31.4}{9.4} & \var{38.2}{8.4} & \var{46.5}{9.0} & \var{63.2}{10.2} & \var{35.6}{8.7} & \var{47.1}{8.1} & \var{55.0}{9.6} & \var{46.9}{10.1} & \var{45.5}{8.9} \\
    & \textit{\makecell{\textbf{GEM3D} \\ [-4pt]{\tiny(ours)}}} & \textbf{\var{44.7}{3.2}} & \textbf{\var{44.3}{3.1}} & \textbf{\var{54.0}{7.2}} & \textbf{\var{67.3}{1.5}} & \textbf{\var{40.0}{5.6}}& \textbf{\var{61.0}{22.1}} & \textbf{\var{65.7}{4.5}} & \textbf{\var{72.0}{13.7}} & \textbf{\var{56.1}{12.1}} \\
    \midrule
    
    \multirowcell{6}[-8pt]{Moderate\\Shift}
    & \textit{PerAct} & \perf{7.7}{9.9}{59.65} & \perf{19.7}{7.2}{30.56} & \perf{14.0}{7.2}{67.94} & \perf{0.3}{0.6}{99.35} & \perf{2.0}{1.7}{92.50} & \perf{12.3}{9.6}{22.92} & \perf{19.7}{7.2}{49.14}& \perf{3.3}{3.2}{75.61}& \perf{9.9}{7.7}{62.21}\\
    & \textit{ManiGaussian} & \perf{15.3}{15.9}{43.91} & \perf{17.0}{12.1}{43.33} & \perf{32.3}{2.3}{26.52} & \perf{30.7}{18.6}{53.76} & \perf{18.3}{2.5}{52.18} & \perf{24.3}{10.1}{20.85} & \perf{30.3}{18.0}{47.09}& \perf{12.0}{4.4}{75.67}& \perf{22.5}{7.9}{45.43}\\
    & \textit{\makecell{ManiGaussian \\ [-2pt]{\tiny(w Distill)}}} & \perf{20.3}{10.3}{34.41} & \perf{23.7}{2.3}{33.01} & \perf{34.0}{9.8}{20.32} & \perf{40.0}{22.6}{37.82} & \perf{21.7}{3.2}{35.64} & \perf{42.3}{7.0}{13.02}& \perf{31.7}{4.5}{41.71}& \perf{21.0}{14.8}{51.16} & \perf{29.3}{8.9}{33.39} \\
    & \textit{DP3} & \perf{13.8}{6.5}{53.5} & \perf{23.7}{7.2}{52.8} & \perf{19.5}{6.8}{58.3} & \perf{28.6}{8.1}{55.1} & \perf{10.2}{5.4}{61.9} & \perf{24.4}{7.1}{56.0} & \perf{15.1}{6.0}{54.8} & \perf{13.9}{5.8}{56.0} & \perf{18.6}{6.6}{55.8} \\
    & \textit{Spatial Forcing} & \perf{15.8}{6.8}{49.7} & \perf{19.7}{7.3}{48.4} & \perf{22.4}{7.1}{51.8} & \perf{31.6}{8.5}{50.0} & \perf{17.2}{6.6}{51.7} & \perf{24.1}{7.4}{48.8} & \perf{23.8}{7.6}{56.7} & \perf{19.3}{7.0}{58.8} & \perf{21.7}{7.3}{52.2} \\
    & \textit{\makecell{\textbf{GEM3D} \\ [-2pt]{\tiny(ours)}}} & \textbf{\perf{33.7}{8.5}{24.62}} & \textbf{\perf{33.0}{5.0}{25.56}} & \textbf{\perf{47.7}{8.4}{11.67}} & \textbf{\perf{48.3}{13.6}{28.22}} & \textbf{\perf{34.3}{4.0}{14.18}}& \textbf{\perf{47.0}{14.8}{22.95}} & \textbf{\perf{42.7}{9.5}{35.02}}& \textbf{\perf{41.7}{13.6}{42.13}}& \textbf{\perf{41.0}{6.5}{25.54}} \\
    \midrule
    \multirowcell{6}[-8pt]{Large\\Shift}
    & \textit{PerAct} & \perf{2.8}{3.4}{85.11} & \perf{15.7}{3.1}{44.69} & \perf{9.7}{4.3}{77.86} & \perf{1.0}{2.0}{98.05} & \perf{3.3}{4.1}{87.51} & \perf{4.8}{3.8}{69.81} & \perf{1.5}{1.9}{96.12}& \perf{1.2}{2.4}{91.44}& \perf{5.0}{5.2}{81.32}\\
    & \textit{ManiGaussian} & \perf{10.3}{14.2}{62.20} & \perf{16.3}{8.4}{45.57} & \perf{27.0}{7.0}{38.64} & \perf{23.7}{16.0}{64.32} & \perf{21.5}{6.9}{43.91} & \perf{16.0}{6.2}{47.88} & \perf{0.0}{0.0}{100.00}& \perf{1.3}{1.8}{97.30}& \perf{14.5}{10.0}{62.49}\\
    & \textit{\makecell{ManiGaussian \\ [-2pt]{\tiny(w Distill)}}} & \perf{21.5}{9.2}{45.16} & \perf{16.3}{6.9}{67.92} & \perf{32.0}{8.6}{29.69} & \perf{33.7}{13.1}{41.97} & \perf{23.2}{12.0}{62.38} & \perf{28.2}{8.0}{53.42} & \perf{0.7}{1.2}{98.77}& \perf{5.3}{7.4}{87.60}& \perf{20.1}{12.0}{60.86}\\
    & \textit{DP3} & \perf{8.7}{4.7}{70.7} & \perf{15.4}{5.8}{69.3} & \perf{13.2}{5.3}{71.8} & \perf{17.8}{6.2}{72.1} & \perf{6.1}{4.1}{77.2} & \perf{14.7}{5.6}{73.5} & \perf{8.4}{4.9}{74.9} & \perf{7.6}{4.5}{75.9} & \perf{11.5}{5.1}{72.8} \\
    & \textit{Spatial Forcing} & \perf{9.6}{5.1}{69.4} & \perf{12.8}{5.9}{66.5} & \perf{14.7}{5.6}{68.4} & \perf{18.5}{6.4}{70.7} & \perf{10.4}{5.0}{70.8} & \perf{15.3}{5.8}{67.5} & \perf{13.7}{5.5}{75.1} & \perf{11.9}{5.2}{74.6} & \perf{13.4}{5.6}{70.6} \\
    & \textit{\makecell{\textbf{GEM3D} \\ [-2pt]{\tiny(ours)}}} & \textbf{\perf{30.2}{7.1}{32.46}} & \textbf{\perf{26.0}{5.8}{41.35}} & \textbf{\perf{45.3}{6.3}{16.11}} & \textbf{\perf{39.0}{13.2}{42.08}} & \textbf{\perf{35.8}{7.7}{10.42}} & \textbf{\perf{40.3}{6.8}{33.88}} & \textbf{\perf{11.0}{15.0}{83.25}}& \textbf{\perf{8.2}{10.0}{88.56}}& \textbf{\perf{29.5}{13.7}{43.53}}\\

    \bottomrule
    \end{tabular}%
    }
    \end{minipage}
\par

%% file: tables/rendering_quality.tex
\par\noindent
\begin{minipage}{\textwidth}
\captionof{table}{
    Comparison of scene reconstruction quality.
    Larger PSNR and SSIM indicate better image fidelity, while lower Chamfer $L_{2}$ indicates higher point cloud geometric accuracy.
}

\small

\setlength{\tabcolsep}{2.5pt} 
\label{tab:rendering_quality}
\begin{tabular*}{\columnwidth}{@{\extracolsep{\fill}}c c c c c} 
\toprule
\textbf{Methods} & \textbf{Tasks} & PSNR$\uparrow$ & SSIM$\uparrow$ & Chamfer $L_{2}$$\downarrow$\\
\midrule
\multirow{4}{*}{\textit{\makecell{ManiGaussian}}}
& \small{\texttt{Open Drawer}} & 16.24 & 0.3926 & 0.04035 \\
& \small{\texttt{Turn Tap}} & 15.84  & 0.3741 & 0.05148 \\
& \small{\texttt{Water Plants}}& 16.07 & 0.3846 & 0.04352 \\
\cmidrule(lr){2-5} 
& \textbf{Avg}& 16.05 & 0.3838 & 0.04511 \\
\midrule 
\multirow{4}{*}[-4pt]{\textit{\makecell{\textbf{GEM3D} \\[-4pt]\scriptsize(ours)}}}
& \small{\texttt{Open Drawer}} & 23.54 & 0.8259 & 0.01541 \\
& \small{\texttt{Turn Tap}} & 22.92  & 0.8123 & 0.01708 \\
& \small{\texttt{Water Plants}}& 22.93 & 0.8025 & 0.01613 \\
\cmidrule(lr){2-5} 
& \textbf{Avg}& \textbf{23.13} & \textbf{0.8136}  & \textbf{0.01621}\\
\bottomrule
\end{tabular*}
\end{minipage}
\par

%% file: tables/multi_task.tex
\begin{table}[!t]

\caption{Multi-task results on four RLBench tasks (SR\%). 
    \textbf{Bold} numbers denote the best performance, and
    $\downarrow$ indicates the relative performance drop compared to the single-task training setting.}
\small
\label{tab:multi_task}
\setlength{\tabcolsep}{4pt}
\begin{tabular*}{\columnwidth}{@{\extracolsep{\fill}} c ccc} 
\toprule
\textbf{Methods}           & Training View          & Moderate Shift      & Large Shift \\
\midrule
\textit{\makecell{PerAct}} 
& \perf{26.1}{17.5}{24.1} 
& \perf{6.4}{4.5}{11.1}
& \makecell{5.6 \scriptsize$\pm$ 3.6 \\ [-0.5ex]\scriptsize(16.1\%$\uparrow$)}  \\
\textit{\makecell{ManiGaussian \\ [-4pt]{\scriptsize(w Distill)}}} 
& \perf{47.1}{5.8}{0.4} 
& \perf{30.7}{8.1}{3.8} 
& \perf{26.5}{7.7}{10.7} \\
\textit{\makecell{\textbf{GEM3D} \\ [-4pt]{\scriptsize(ours)}}} 
& \textbf{\makecell{52.9 \scriptsize$\pm$ 6.4 \\ [-0.5ex]\scriptsize(1.7\%$\uparrow$)}} 
& \textbf{\perf{41.2}{9.6}{6.9}} 
& \textbf{\perf{36.4}{9.0}{3.7}} \\
\bottomrule
\end{tabular*}
\end{table}

%% file: tables/multi_task_result.tex
\begin{table*}[!t]
    \centering
    \caption{
        Results of multi-task view generalization on RLBench tasks.
        We report the success rate (mean $\pm$ std \%) in the multi-task setting.
        \textbf{Bold} numbers denote the best performance.
        The percentage in parentheses indicates the relative performance change compared to the single-task baseline ($\frac{\text{Multi}-\text{Single}}{\text{Single}}$).
    }
    \label{tab:multi_task_result}
    \small
    \setlength{\tabcolsep}{4pt}
    \begin{tabular*}{\textwidth}{@{\extracolsep{\fill}} c c ccccc}
    \toprule
    & \textbf{Methods} & \texttt{\makecell{Open \\ Jar}} & \texttt{\makecell{Meat on \\ Grill}} & \texttt{\makecell{Turn \\ Tap}} & \texttt{\makecell{Phone on \\ Base}} & \textbf{Avg.}\\
    \midrule
    \multirowcell{4}[-6pt]{Moderate\\Shift} 
    & \textit{PerAct} 
        & \perfs{4.3}{4.0}{115.00}{\uparrow} 
        & \perfs{2.0}{2.6}{83.74}{\downarrow} 
        & \perfs{15.7}{5.0}{12.14}{\uparrow} 
        & \perfs{3.7}{6.4}{1133.33}{\uparrow} 
        & \perfs{6.4}{4.5}{11.11}{\downarrow}\\
    & \textit{\makecell{ManiGaussian \\ [-4pt]{\scriptsize(w Distill)}}}
        & \perfs{21.0}{7.9}{3.23}{\downarrow} 
        & \perfs{28.7}{10.1}{32.15}{\downarrow} 
        & \perfs{44.3}{9.1}{30.29}{\uparrow} 
        & \perfs{28.7}{5.1}{28.25}{\downarrow} 
        & \perfs{30.7}{8.1}{11.01}{\downarrow}\\
    & \textit{\makecell{\textbf{GEM3D} \\ [-4pt]{\scriptsize(ours)}}} 
        & \textbf{\perfs{27.0}{13.0}{21.28}{\downarrow}} 
        & \textbf{\perfs{38.7}{14.5}{17.66}{\downarrow}} 
        & \textbf{\perfs{59.0}{4.6}{23.69}{\uparrow}} 
        & \textbf{\perfs{40.0}{6.2}{17.18}{\downarrow}} 
        & \textbf{\perfs{41.2}{9.6}{7.00}{\downarrow}} \\
    
    \midrule
    \multirowcell{4}[-6pt]{Large\\Shift} 
    & \textit{PerAct} 
        & \perfs{5.3}{5.2}{60.61}{\uparrow} 
        & \perfs{1.2}{1.9}{75.00}{\downarrow} 
        & \perfs{14.5}{5.4}{49.48}{\uparrow} 
        & \perfs{1.5}{1.8}{50.00}{\uparrow} 
        & \perfs{5.6}{3.6}{19.15}{\uparrow}\\
    & \textit{\makecell{ManiGaussian \\ [-4pt]{\scriptsize(w Distill)}}}
        & \perfs{22.2}{7.2}{16.23}{\downarrow} 
        & \perfs{26.3}{9.3}{6.05}{\uparrow} 
        & \perfs{36.0}{10.2}{20.00}{\uparrow} 
        & \perfs{21.3}{5.7}{42.90}{\downarrow} 
        & \perfs{26.5}{8.1}{10.77}{\downarrow}\\
    & \textit{\makecell{\textbf{GEM3D} \\ [-4pt]{\scriptsize(ours)}}} 
        & \textbf{\perfs{27.0}{10.7}{9.40}{\downarrow}} 
        & \textbf{\perfs{31.3}{7.4}{15.86}{\downarrow}} 
        & \textbf{\perfs{47.0}{12.6}{3.75}{\uparrow}} 
        & \textbf{\perfs{40.3}{5.4}{3.33}{\uparrow}} 
        & \textbf{\perfs{36.4}{9.0}{3.70}{\downarrow}}\\
    
    \bottomrule
    \end{tabular*}
    \end{table*}